\documentclass[runningheads]{llncs}
\usepackage{subcaption}
\usepackage[T1]{fontenc}
\usepackage{graphicx}
\usepackage{amsmath}
\usepackage{amssymb}
\usepackage{algorithm}
\usepackage{algpseudocode}
\usepackage{xcolor}
\usepackage{hyperref}
\usepackage{color}

\begin{document}
\title{Concept-based Visual Counterfactual Explanations with Diffusion Models}
\author{Yassine Oueslati\inst{1}, Daniil Kirilenko\inst{1}, Martin Gjoreski\inst{1}, Marc Langheinrich\inst{1}}
\authorrunning{Y. Oueslati}
\institute{Università della Svizzera italiana, Lugano, Switzerland\\
\email{name.surname@usi.ch}}
\maketitle
\begin{abstract}


Visual counterfactual explanations aim to answer "what minimal change to this image would flip the model's prediction?", and are increasingly important as vision models are deployed in safety-critical domains (e.g., medicine). Existing diffusion-based methods can produce realistic edits, but they rely on external classifiers that must work reliably on noisy images, which makes them fragile and hard to deploy for robust explanations. We introduce C-VCE, a new diffusion framework that builds the classifier directly into the generative model via a concept bottleneck layer, so that counterfactuals are guided by human-interpretable features (concepts) instead of a separate noise-robust classifier that works with pixel-level edits. Our model lets users to toggle on/off semantic concepts during sampling, then minimally adjusts relevant image regions, while preserving the rest of the image, respecting feature correlations. To keep edits small and controlled, we add a simple probabilistic regularizer that balances "change the prediction" against "stay close to the original", plus a gradient-based mask that confines modifications to the most relevant regions. On benchmarks such as CelebA, C-VCE matches or improves flip rates while producing counterfactuals that are visually closer to the input and less distorted than baselines that depend on separate noisy-image classifiers. These properties make C-VCE a practical tool for vision systems where users need concrete "what-if" images without having to trust an additional, noise-robust classifier. More broadly, our results suggest that exposing and controlling an internal concept layer is a promising way to make powerful generative models easier to understand and safer to use.\\ 
\textbf{Code}: \href{https://github.com/yassine2331/ex_diffuser}{https://github.com/yassine2331/ex\_diffuser}

\end{abstract}

\section{Introduction}
\label{sec:introduction}


As artificial intelligence (AI) systems become increasingly integrated into high-stakes domains such as medical diagnostics \cite{exp:healthcare:saraswat2022explainable,medical:tjoa2020survey,medical:krishnan2022self}, and autonomous systems \cite{exp:car:kuznietsov2024explainable,car:atakishiyev2024explainable}, the demand for transparency has become paramount. Many powerful deep learning models operate as "black boxes," delivering highly accurate predictions without revealing the underlying reasoning behind their decisions \cite{expl:das2020opportunitieschallengesexplainableartificial,blackbox:rai2020explainable}. This opacity erodes trust and hinders accountability, especially in critical scenarios where unexplained errors carry severe consequences. Consequently, research has intensified into Explainable AI (XAI)\cite{explain:dwivedi2023explainable} and the development of post-hoc explanation methods—tools designed to interpret a model’s logic after it has been trained. This shift is further reinforced by legal frameworks like the GDPR’s "right to explanation" \cite{right:to:explain:Goodman_2017}, which underscores the regulatory necessity for clarity in automated decision-making.


Moving from general explainability to specific methodologies, this work investigates Visual Counterfactual Explanations (VCEs) \cite{VCE:goyal2019counterfactual}. A VCE answers a simple yet profound question: \textit{"What is the smallest change that can be made to an input image to produce a different decision from the model?"} \cite{counterfactual:molnar2020interpretable}. Unlike attribution-based methods (e.g., saliency maps, Grad-Cam)\cite{attribusion:method:adebayo2018sanity,Selvaraju_2019} that highlight which pixels were influential, VCEs provide a contrastive perspective. For instance, in an MRI-based tumor classification, a VCE helps a doctor determine if a diagnosis was driven by the shape, size, or texture of a specific region \cite{exp:healthcare:saraswat2022explainable,GAN:VCE:medical:mertes2022ganterfactual}. Formally, VCEs are typically defined by three core properties: they must be valid (the model's prediction actually changes), plausible (the modified image remains realistic), and maintain proximity (the change is minimal) \cite{counter:factual:DBLP:journals/datamine/Guidotti24}.


There is a fine line between generating VCEs and Adversarial Examples (AEs)\cite{adv-example:goodfellow2015explainingharnessingadversarialexamples}. While both involve changing a model's prediction through minimal perturbations, they serve fundamentally different objectives \cite{AE:VCE:freiesleben2022intriguing}. The goal of an AE is to "fool" the model using small, often invisible changes that lack semantic meaning. To ensure that VCEs provide human-understandable insights rather than mere adversarial noise, one could train robust models \cite{rebust:augustin2020adversarialrobustnessinoutdistribution}; however, this significantly increases training complexity \cite{no:robust:prach2025intriguingpropertiesrobustclassification}.

Consequently, research has shifted toward utilizing generative models \linebreak \cite{GAN:VCE:medical:mertes2022ganterfactual,CFE:1:jeanneret2022diffusion,semantic:gan:jacob2022steexsteeringcounterfactualexplanations,semantic:gan:samadi2023safesaliencyawarecounterfactualexplanations,semantic:gan:sauer2021counterfactualgenerativenetworks,cycle:gan:ghandeharioun2022dissectdisentangledsimultaneousexplanations,cycle:gan:khorram2022cycleconsistentcounterfactualslatenttransformations,style:gan:lang2021explainingstyletraininggan}, which excel at addressing this gap by ensuring that edits remain visually plausible and aligned with the data distribution \cite{daniil:kirilenko2024generative}. In this work, we focus on Denoising Diffusion Probabilistic Models (DDPMs) \cite{DDPM:ho2020denoisingdiffusionprobabilisticmodels}, which have proven to be the state-of-the-art in image synthesis, quality, and editing tasks \cite{DDPM:classifier:dhariwal2021diffusionmodelsbeatgans}. By learning to reverse a progressive noise process, DDPMs allow us to generate high-fidelity counterfactuals that are both semantically meaningful and structurally sound\cite{CFE:2:augustin2022diffusionvisualcounterfactualexplanations,CFE:1:jeanneret2022diffusion}.

Since generating a CE requires flipping models' output, we must condition the DDPM process towards the target output. This is typically achieved via (i) classifier guidance\cite{DDPM:classifier:dhariwal2021diffusionmodelsbeatgans} or (ii) classifier-free guidance\cite{DDPM:classifier-free:ho2022classifierfreediffusionguidance}. Related work usually employs the former, which guides the reverse diffusion process using an external classifier. However, this approach has a limitation: it requires the external classifier to be robust to noise, as it must operate on intermediate noisy samples \cite{rebust:santurkar2019imagesynthesissinglerobust}. In contrast, classifier-free guidance integrates the condition directly into the diffusion model during training on conditioned inputs, avoiding the need for an external noise-robust classifier entirely.


Among conditioning approaches \cite{cond:ddpm:panagiotakopoulos2025conditional}, we explore Concept Bottleneck Models (CBMs), which introduce an explicit, human-aligned concept layer between the input and the prediction \cite{CBM:koh2020conceptbottleneckmodels}. This two-stage architecture—predicting semantic concepts before the final classification—enables direct semantic intervention \cite{CBM:DDPM:ismail2024concept}, where adjusting the "bottleneck" allows us to steer the model toward a different outcome.

While recent frameworks such as CF-CBMs have begun generating counterfactuals within this bottleneck \cite{dominici2025counterfactualconceptbottleneckmodels,dominici2024causal}, these approaches have not yet been integrated with DDPMs. Conversely, although hybrid Concept Bottleneck Generative Models have successfully coupled CBMs with DDPMs to improve generative control \cite{CBM:DDPM:ismail2024concept}, their primary objective has been high-fidelity image synthesis rather than the generation of counterfactual explanations.


The final hurdle in VCE generation is localization, i.e., ensuring that the model only modifies relevant semantic regions while keeping the rest of the image untouched. Most existing image editing methods are computationally inefficient for this task \cite{CFE:1:jeanneret2022diffusion,CFE:ADVER:Jeanneret_2023_CVPR}. For instance, some approaches require extensive reverse diffusion steps to combat noise \cite{CFE:1:jeanneret2022diffusion}, while others, such as ACE, utilize a two-stage process where an initial diffusion pass is performed to extract a mask, followed by a second pass to apply the localized edit \cite{CFE:ADVER:Jeanneret_2023_CVPR}. These multi-pass requirements, along with methods that rely on comparing image estimates with the original input, make them impractical for the fast, automated generation of counterfactual explanations.


We introduce C-VCE (Concept-based Visual Counterfactual Explanations), a framework that bridges these gaps by utilizing CBMs to guide the diffusion process toward valid, visual counterfactuals that are both semantically grounded and visually plausible. We address these challenges through:
\begin{itemize}
    \item \textbf{First CBM-Integrated Diffusion Pipeline:} To the best of our knowledge, C-VCE is the first method to embed a CBM directly into the generative U-Net. By bridging the gap between bottleneck-based reasoning and image generation, we allow users to "steer" counterfactuals using human-understandable concepts.

    \item  \textbf{Score-based Proximity via Product-of-Experts:} We propose a novel proximity regularizer that treats "validity" and "proximity" as two competing experts. This provides a mathematically grounded way to balance the two without needing expensive clean-image approximations at every step.

    \item  \textbf{Integrated Gradient-Based Masking:} Unlike methods that require a separate initial pass to extract a mask, our framework computes a dynamic, gradient-derived mask on-the-fly during the diffusion process. This enables precisely localized edits in a single reverse diffusion pass, preventing reconstruction collapse while significantly reducing computational overhead.
\end{itemize}

\section{Background and Related Work}
\label{sec:background}

This section provides technical background used throughout the paper. We first define CEs and discuss why enforcing visual plausibility is challenging in high-dimensional image spaces. We then review diffusion models and conditioning mechanisms, and finally summarize CBMs as an interpretable interface for controllable counterfactual generation.

\subsection{CEs and the Visual Plausibility Challenge}

\begin{definition}
A counterfactual explanation of a prediction describes the smallest change to the feature values that changes the prediction to a predefined output  \cite{counterfactual:molnar2020interpretable} 
\end{definition}

 Formally, let $f_\theta$ be a model that, for a given input $\hat{x}$, outputs a decision $\hat{y} = f_\theta(\hat{x})$. A counterfactual explanation is an alternative instance $x'$ such that $f_\theta(x') = y'$, where $y'$ is a desired target outcome. In practice, $x'$ is often sought to be as close as possible to $\hat{x}$ under a chosen distance metric. This is typically framed as a constrained optimization problem where we seek to minimize a distance metric $D(x, \hat{x})$ subject to a change in the model's output \cite{counter:factual:basics:wachter2018counterfactualexplanationsopeningblack}. In practice, this is often relaxed into a penalty function:


To satisfy the counterfactual objective, we define a loss function $\mathcal{L}_{CE}$ for a candidate explanation $x'$ relative to the original input $\hat{x}$:
\begin{equation}
    \mathcal{L}_{CE}(x') = \lambda \cdot \mathcal{L}_{pred}(f_\theta(x'), y') + D(x', \hat{x})
\end{equation}
where $\mathcal{L}_{pred}$ measures how far the prediction is from the desired one, $y'$ is the desired target class, $f_\theta$ is the classifier with parameters $\theta$, and $D(\cdot)$ is a distance metric. This formulation allows for the use of optimization-based strategies to produce new counterfactuals by balancing two core properties: \textbf{validity}, ensuring the model's prediction changes to the desired target ($f_\theta(x') = y'$), and \textbf{proximity}, which minimizes the distance $D(x', \hat{x})$ to ensure the counterfactual remains as close as possible to the original input. However, for an explanation to be practically useful, it must also satisfy a \textbf{plausibility} criterion. This property requires the introduced modifications to be plausible from a \textit{human perspective}, which is technically interpreted as ensuring the instance $x'$ lies within the data manifold and does not appear as an outlier.



In high-dimensional settings, such as image data where $x \in \mathbb{R}^{C \times H \times W}$ represents pixel values across channels, height, and width, identifying the smallest possible perturbation that alters a classifier's decision faces a critical challenge. Modern deep neural networks are highly susceptible to adversarial examples—small perturbations that change the predicted class while leaving the image visually unchanged for humans \cite{curse:szegedy2014intriguingpropertiesneuralnetworks,adv-example:goodfellow2015explainingharnessingadversarialexamples}. 

Mathematically, generating a VCE can inadvertently collapse into an adversarial attack \cite{AE:VCE:freiesleben2022intriguing}, such as the Fast Gradient Sign Method (FGSM)\cite{adv-example:goodfellow2015explainingharnessingadversarialexamples}:
\begin{equation}
    x_{adv} = \hat{x} + \eta \cdot \operatorname{sign}(\nabla_x \mathcal{L}_{pred}(f_\theta(\hat{x}), y'))
\end{equation}
where $\eta$ represents the perturbation step size. Unlike true counterfactuals, these adversarial instances lack semantic plausibility because they exploit non-robust features of the model rather than altering human-understandable concepts.

The conflict arises because an adversarial image $x_{adv}$ often satisfies the mathematical requirements for both validity and proximity. However, because these perturbations exploit the model's vulnerabilities rather than its semantic understanding, they fundamentally violate the property of plausibility \cite{daniil:kirilenko2024generative,AE:VCE:freiesleben2022intriguing}. The resulting counterfactual is often untrustable, as it fails to highlight the actual features that would need to change in order to reach a different decision.

To address this, various methods attempt to enforce plausibility by penalizing deviations from the data manifold $\mathcal{M}_X$ \cite{counter:factual:basics:wachter2018counterfactualexplanationsopeningblack} or by leveraging adversarially robust models \cite{VCE:NO:GENERATIVEMODEL:chang2021robustclassificationmodelcounterfactual}. While robust models produce more semantically coherent VCEs, they are often significantly less accurate than their standard counterparts \cite{no:robust:prach2025intriguingpropertiesrobustclassification}. More broadly, generative models provide a principled framework for capturing the data distribution and producing realistic, semantically meaningful counterfactuals \cite{daniil:kirilenko2024generative}.

\subsection{Generative Modeling with Diffusion}

\textbf{Generative Models.} To ensure counterfactuals look realistic (plausibility), early methods used GANs and VAEs to edit images in a compressed latent space~\cite{condGAN:Samangouei_2018_ECCV,vae:issue:zhao2017toward}. However, GANs are often unstable to train, and VAEs tend to produce blurry results~\cite{GAN:original:goodfellow2014generativeadversarialnetworks}. Recently, Denoising Diffusion Probabilistic Models (DDPMs) have become the standard because they offer high-quality image generation with stable training~\cite{DDPM:classifier:dhariwal2021diffusionmodelsbeatgans}.

Formally, diffusion models work by reversing a process that gradually adds noise to an image. We define a forward process that adds Gaussian noise to a clean sample $\mathbf{x}_0$ over time steps $t$, eventually turning it into pure noise $\mathbf{x}_T$. A key property is that we can sample any noisy state $\mathbf{x}_t$ directly:

\begin{equation}
\mathbf{x}_t \sim q(\mathbf{x}_t| \mathbf{x}_0) = \mathcal{N}(\mathbf{x}_t; \sqrt{\bar{\alpha}_t}\mathbf{x}_0 , (1 - \bar{\alpha}_t)\mathbf{I})
\end{equation}

\begin{equation}
\mathbf{x}_t = \sqrt{\bar{\alpha}_t} \mathbf{x}_0 + \sqrt{1 - \bar{\alpha}_t} \boldsymbol{\epsilon}, \quad \boldsymbol{\epsilon} \sim \mathcal{N}(0, \mathbf{I})
\end{equation}

To generate new data, a Unet \cite{unet:ronneberger2015unetconvolutionalnetworksbiomedical} is trained to predict the noise  $\boldsymbol{\epsilon}_\theta$ using the simplified loss from Ho et al.~\cite{DDPM:ho2020denoisingdiffusionprobabilisticmodels}. Once trained, we generate new samples by iteratively removing the noise using the predicted $\boldsymbol{\epsilon}_\theta$:
\begin{equation}
\mathbf{x}_{t-1} = \frac{1}{\sqrt{\alpha_t}} \left( \mathbf{x}_t - \frac{1 - \alpha_t}{\sqrt{1 - \bar{\alpha}_t}} \boldsymbol{\epsilon}_\theta(\mathbf{x}_t, t) \right) + \sigma_t \mathbf{z}
\end{equation}
where $\mathbf{z} \sim \mathcal{N}(0, \mathbf{I})$ adds stochasticity to the reverse process.

\textbf{Score-Based Modeling and Guidance.} While DDPMs are implemented as noise predictors, they are theoretically grounded in score-based generative modeling~\cite{score:song2020score}. In this framework, we aim to estimate the \textit{score function} $\mathbf{s}_\theta(\mathbf{x}_t, t)$, defined as the gradient of the log-probability density $\mathbf{s}_\theta(\mathbf{x}_t, t) \approx \nabla_{\mathbf{x}_t} \log p(\mathbf{x}_t)$. Intuitively, this function represents a vector field pointing toward regions of maximum data likelihood. By following these gradients (via Langevin dynamics), we can traverse from a low-density region (noise) to a high-density region (clean data).

The crucial utility of this perspective is the exact relationship between the score and the noise predictor:
\begin{equation}
\boldsymbol{\epsilon}_\theta(\mathbf{x}_t, t) = -\sqrt{1 - \bar{\alpha}_t} \, \mathbf{s}_\theta(\mathbf{x}_t, t)
\end{equation}
This relationship serves as a powerful translation mechanism: it allows us to manipulate probabilities using standard rules (like Bayes' theorem) and then immediately translate those changes into noise prediction updates, bypassing complex diffusion derivations. For example, to steer generation toward a class $y$, we apply Bayes' rule to the score: $\nabla_{\mathbf{x}} \log p(\mathbf{x} | y) = \nabla_{\mathbf{x}} \log p(\mathbf{x}) + \nabla_{\mathbf{x}} \log p(y | \mathbf{x})$. where $p(y | \mathbf{x})$ represents the output of a classifier trained on noisy data. By translating this equation back into the DDPM framework using the relationship above, we obtain \textbf{Classifier Guidance}, where the unconditional noise is shifted by a classifier's gradient~\cite{DDPM:classifier:dhariwal2021diffusionmodelsbeatgans}:
\begin{equation}
\boldsymbol{\hat{\epsilon}}(\mathbf{x}_t, t, y) = \boldsymbol{\epsilon}_\theta(\mathbf{x}_t, t) - \sqrt{1-\bar{\alpha}_t} \nabla_{\mathbf{x}_t} \log p(y | \mathbf{x}_t)
\end{equation}

Similarly, we can apply this logic to \textbf{Classifier-Free Guidance}, where the conditional score is extrapolated from the unconditional score. Translated into noise predictions, this yields a formula that maximizes conditional likelihood without requiring an external classifier~\cite{DDPM:classifier-free:ho2022classifierfreediffusionguidance}:
\begin{equation}
\boldsymbol{\hat{\epsilon}}(\mathbf{x}_t,t,y) = (1+w)\boldsymbol{\epsilon}_\theta (\mathbf{x}_{t},t,y) - w\boldsymbol{\epsilon}_\theta (\mathbf{x}_{t},t, \emptyset)
\end{equation}

where $\boldsymbol{\epsilon}_\theta (\mathbf{x}_{t},t,y)$ is the noise predicted given the target class $y$, and $\boldsymbol{\epsilon}_\theta (\mathbf{x}_{t},t, \emptyset)$ is the unconditional noise prediction where the class information is omitted.

This formulation ensures the counterfactual remains plausible (by following the data manifold via $\boldsymbol{\epsilon}_\theta(\cdot, \emptyset)$) while satisfying the validity constraint (by moving toward the target class via the guidance term).

\textbf{Diffusion-Driven Visual Counterfactuals.} 
Leveraging the guidance mechanisms mentioned above, 
researchers have applied diffusion models to VCEs. 
The standard approach follows a ``perturb-and-denoise'' process: an input image $\mathbf{x}_0$ is first corrupted with noise up to a timestep $\tau$ to erase specific details. 
The model then performs a guided reverse process starting from $\mathbf{x}_\tau$ to create a new sample $\mathbf{x}'_0$. This new sample aims to meet the counterfactual target $y'$ while retaining the structural fidelity of the original input\cite{CFE:1:jeanneret2022diffusion,CFE:2:augustin2022diffusionvisualcounterfactualexplanations,CFE:ADVER:Jeanneret_2023_CVPR}.

However, guiding this reverse process is challenging because standard classifiers are not robust to the noisy intermediate steps typical of diffusion models~\cite{DDPM:classifier:dhariwal2021diffusionmodelsbeatgans}. 
Methods differ mainly in how they solve this problem. 
One category uses \textbf{explicit guidance}. 
For example, \textbf{DiME}~\cite{CFE:1:jeanneret2022diffusion} employs a loss-guided approach rather than simple classifier guidance. 
At each step, it first estimates a clean image using unconditional diffusion, then calculates a composite loss (combining classification and perceptual similarity) on this clean estimate to steer the noisy sample. 
While this avoids the need for robust classifiers, DiME \cite{CFE:1:jeanneret2022diffusion} performs the clean estimate calculation at every step, requiring around 1800 steps per counterfactual. 
To address this computational burden, methods like \textbf{Fast-DiME}~\cite{CF:FAST:weng2024fastdiffusionbasedcounterfactualsshortcut} have been proposed to speed up the process by reducing the number of required gradient evaluations. 
\textbf{DVCE}~\cite{CFE:2:augustin2022diffusionvisualcounterfactualexplanations} takes a different approach by directly modifying the reverse diffusion transition during a single sampling trajectory. It combines classifier gradients with distance regularization and uses cone projection to stabilize the guidance, allowing it to work with non-robust classifiers without the heavy overhead of repeated reconstruction. Extending this concept, \textbf{Latent-DVCE (L-DVCE)} adapts this framework to operate within the latent space of latent diffusion models rather than pixel space.

Other works try to further reduce these costs by optimizing the sampling schedule or operating in a compressed latent space to improve efficiency~\cite{CF:LATENT:DDPM:Luu_2025}. 
A separate category uses \textbf{textual inversion} or prompt mixing to change image content by adjusting conditioning embeddings instead of using direct gradients~\cite{CF:COCO:le2023cococounterfactualsautomaticallyconstructedcounterfactual}.

While these methods effectively generate counterfactuals, they rely on standard DDPMs which operate as ``black boxes'' regarding their internal semantic decisions. 
This raises a critical question: can we add a layer of interpretability on top of this generative process? 
Ideally, when generating a counterfactual, we should be able to inspect \textit{what} specific concepts drove the model to that conclusion. 
Furthermore, we should be able to \textit{intervene} on these concepts during inference to control the explanation explicitly. 
To enable this level of transparency and control, the next section introduces CBMs.

\subsection{Controllable Generation via Concept Bottlenecks}

\textbf{CBMs}~\cite{CBM:koh2020conceptbottleneckmodels} are a type of interpretable model designed to make the decision process transparent by introducing an intermediate step that aligns with human understanding. Instead of learning a direct, opaque mapping from an input $x$ to an output $y$, CBMs break the process down: first, they predict a set of human-defined concepts $\hat{c}$ from the input, and then they use these concepts to predict the final output $\hat{y}$. Formally, a CBM consists of two functions:
\begin{equation}
\begin{aligned}
g : \mathbb{R}^d \rightarrow \mathbb{R}^k&, \quad f : \mathbb{R}^k \rightarrow \mathbb{R}^o, \quad \hat{y} = f(g(x)),\\
x \to \hat{c} \quad&, \quad \quad \quad  \hat{c} \to \hat{y} ,
\end{aligned}
\end{equation}

where $d$ represents the dimensionality of the input space, $k$ denotes the number of human-defined concepts in the bottleneck layer and $o$ is the output dimension. 

\textbf{Training and Intervention.} The training of CBMs typically involves a multi-task objective. In a joint bottleneck regime, the model minimizes a combined loss that balances task accuracy $\mathcal{L}_Y$ and concept prediction accuracy $\mathcal{L}_{C_j}$ for each concept $j$:
\begin{equation}
    \min_{f,g} \sum_i \left[ \mathcal{L}_Y(f(g(x^{(i)})), y^{(i)}) + \lambda \sum_j \mathcal{L}_{C_j}(g_j(x^{(i)}), c_j^{(i)}) \right]
\end{equation}
A primary advantage of this structure is the capacity for \textbf{test-time intervention}. Users can manually override a predicted concept $\hat{c}_j$ with a corrected value and observe the change in the final prediction $\hat{y}$. This mechanism enables a structured form of counterfactual reasoning—asking ``If this specific concept were different, how would the model's decision change?''—allowing domain experts to correct model reasoning without retraining~\cite{CBM:koh2020conceptbottleneckmodels}.

\textbf{CBM-DDPM Integration.} To leverage these properties for generative tasks, CBMs have been integrated into the DDPM framework by conditioning the denoising network on the concept space~\cite{CBM:DDPM:ismail2024concept}. In this architecture, the standard U-Net bottleneck is replaced with a CBM module. The network $\epsilon_\theta$ is thus composed of an encoder $e$, the concept bottleneck components, and a decoder $g$.

Formally, the encoder extracts a pre-concept embedding and skip features: $e(\mathbf{x}_t) \to (\mathbf{h}_{\text{pre}}, \mathbf{s})$. The CBM module then predicts concept probabilities via a classifier head $f_c$ and constructs a post-concept embedding $\mathbf{h}_{\text{post}}$ using an embedding constructor $f_e$. Finally, the decoder predicts the noise from this concept-aware embedding: $\boldsymbol{\epsilon} = g(\mathbf{h}_{\text{post}}, \mathbf{s})$.

This design enables two distinct modes of operation:
\begin{itemize}
    \item \textbf{Unconditional Generation:} The model infers concepts directly from the noisy input to guide the denoising step.
    \begin{equation}
    c = f_c(\mathbf{h}_{\text{pre}}), \quad \mathbf{h}_{\text{post}} = f_e(c, \mathbf{h}_{\text{pre}}), \quad \epsilon_\theta(\mathbf{x}_t, t, \emptyset) = g(\mathbf{h}_{\text{post}}, \mathbf{s})
    \end{equation}
    \item \textbf{Conditional Generation:} An external target concept vector $c'$ is provided to intervene on the generation process.
    \begin{equation}
    \mathbf{h}_{\text{post}} = f_e(c', \mathbf{h}_{\text{pre}}), \quad \epsilon_\theta(\mathbf{x}_t, t, c') = g(\mathbf{h}_{\text{post}}, \mathbf{s})
    \end{equation}
\end{itemize}
This mechanism allows the user to steer the reverse diffusion process by injecting specific semantic targets $c'$ directly into the bottleneck, satisfying the validity constraint through interpretable interventions.

\section{Methodology}
\label{sec:methodology}



\begin{figure}[h!]
    \centering
    \includegraphics[width=1\textwidth]{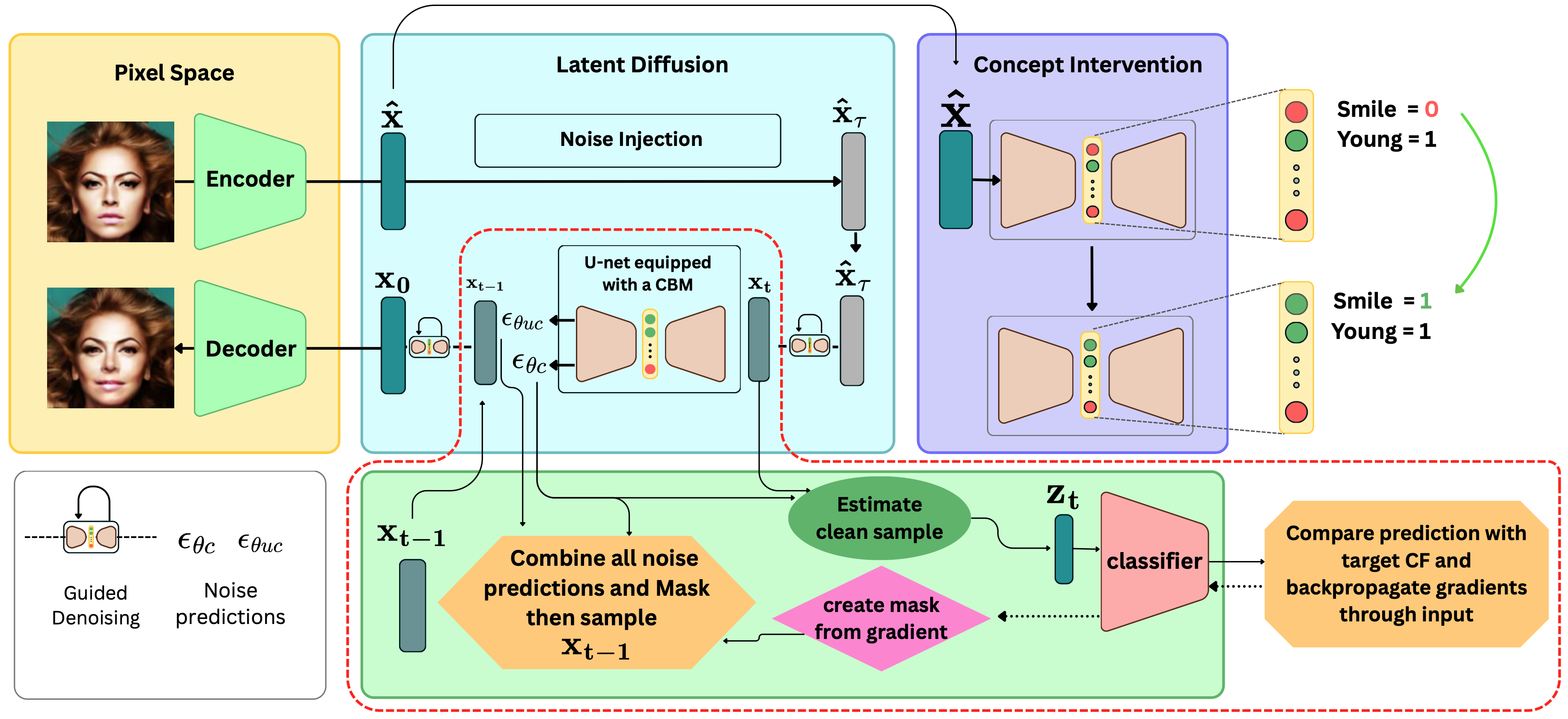}
    \caption{\textbf{Overview of the C-VCE Framework.} The process begins with encoding the original image into latent space. Within the \textbf{Latent Diffusion} phase, a U-Net equipped with a CBM performs guided denoising. The \textbf{Concept Intervention} module allows for targeted attribute shifts (e.g., Smile $0 \to 1$). Simultaneously, the framework estimates clean samples to compute gradients via an external classifier, creating a dynamic \textbf{gradient-derived mask}. Finally, noise predictions and the mask are combined to iteratively sample the previous time step until a clean latent counterfactual is reached, which is then decoded back into pixel space.}
    \label{fig:cvce_overview}
\end{figure}

This section describes the proposed \textbf{C-VCE (Concept-based Visual Counterfactual Explanations)} framework, illustrated in Fig.~\ref{fig:cvce_overview}. The framework is designed to generate VCEs that maintain a balance between \textbf{validity}, \textbf{plausibility}, and \textbf{proximity}. We first introduce the overall diffusion-based formulation and detail the integration of a CBM directly into the U-Net bottleneck to enable explicit concept-level control. We then present our product-of-experts regularizer, which enforces proximity by softly anchoring the reverse diffusion trajectory to the original input. Finally, we detail the resulting training objective and the counterfactual sampling algorithm that enables localized edits in a single pass.

\subsection{Concept-level Counterfactual Interventions}


Our framework achieves semantic validity by embedding a CBM directly into the latent bottleneck of the denoising U-Net. This architecture allows the model to function simultaneously as a high-fidelity generator and an internal concept predictor.

\paragraph{Latent Encoding.}
We define the input query image as $I \in \mathbb{R}^{3 \times H \times W}$. To operate within a computationally efficient manifold, we first map this image to a lower-dimensional latent representation $x_0 = \text{ENC}(I)$ using a pre-trained VAE encoder. During the reverse diffusion process, the U-Net $\epsilon_\theta$ processes the noisy latent $x_t$ at each timestep $t$.

\paragraph{The Concept Bottleneck.}
We decompose the denoising network $\epsilon_\theta$ into three functional components: a latent encoder $e$, a CBM module (consisting of a classifier head $f_c$ and an embedding constructor $f_e$), and a latent decoder $g$. The process begins by extracting a pre-concept embedding $h_{\text{pre}}$ and a set of skip-connection features $s$ from the noisy latent:
\begin{equation}
    (h_{\text{pre}}, s) = e(x_t, t)
\end{equation}

The CBM module then maps $h_{\text{pre}}$ into a human-interpretable concept space. The classifier head $f_c$ predicts a vector of concept probabilities $c \in [0, 1]^k$, where each dimension represents a specific semantic attribute (e.g., \textit{"Smiling"} or \textit{"Young"}) defined by the dataset:
\begin{equation}
    c = \sigma(f_c(h_{\text{pre}}))
\end{equation}

\paragraph{Semantic Intervention for Validity.}
To generate a counterfactual, we perform a targeted intervention on the predicted concept vector. We replace the inferred concepts $c$ with a target concept vector $c' \in \{0, 1\}^k$, where specific bits are flipped to satisfy the counterfactual condition (e.g., setting the 'Smile' attribute to 1). 

The embedding constructor $f_e$ then fuses this target vector $c'$ back into the latent pipeline to create a post-concept embedding $h_{\text{post}} = f_e(c', h_{\text{pre}})$. The final noise prediction is generated by the decoder $g$:
\begin{equation}
    \epsilon_\theta(x_t, t, c') = g(h_{\text{post}}, s)
\end{equation}

By conditioning the denoising process on $c'$ at the bottleneck level, C-VCE ensures that the generated counterfactual is semantically grounded in human-interpretable features while maintaining structural consistency through the skip features $s$.


\paragraph{Classifier-Free Guidance.}
To amplify semantic control, we utilize classifier-free guidance (CFG) to bias the reverse sampling trajectory by interpolating between conditional and unconditional noise predictions \cite{DDPM:classifier-free:ho2022classifierfreediffusionguidance}. This approach further steers the model toward the target concept manifold. The guided noise prediction $\epsilon^{(w)}$ is computed as:
\begin{equation}
    \epsilon^{(w)}(x_t, t, c') = (1+w)\epsilon_\theta(x_t, t, c') - w\epsilon_\theta(x_t, t, \emptyset)
\end{equation}
where $w \in \mathbb{R}^+$ denotes the guidance scale, $c'$ is the target counterfactual concept vector, and $\emptyset$ represents the null condition—implemented here as the unconditional mode where the model infers concepts directly from the noisy latent $x_t$ rather than through an external intervention. 

Higher values of $w$ concentrate the probability mass into regions of the data distribution most consistent with $c'$. By anchoring the intervention within the CBM bottleneck, the framework ensures the generative process remains directed toward semantically meaningful targets while preserving the high-fidelity reconstruction capabilities of the diffusion backbone.

\subsection{Proximity via Product-of-Experts}



To satisfy the proximity constraint, we introduce a probabilistic regularizer based on a \textbf{product-of-experts} formulation. While the CBM-guided diffusion trajectory ensures the output is semantically valid, it does not inherently guarantee that the image remains close to the original input $\hat{x}$. To enforce this, we treat the generation process as a combination of two independent "experts": one focused on validity and the other on proximity.

\begin{figure}[h!]
    \centering
    \includegraphics[width=1.0\textwidth]{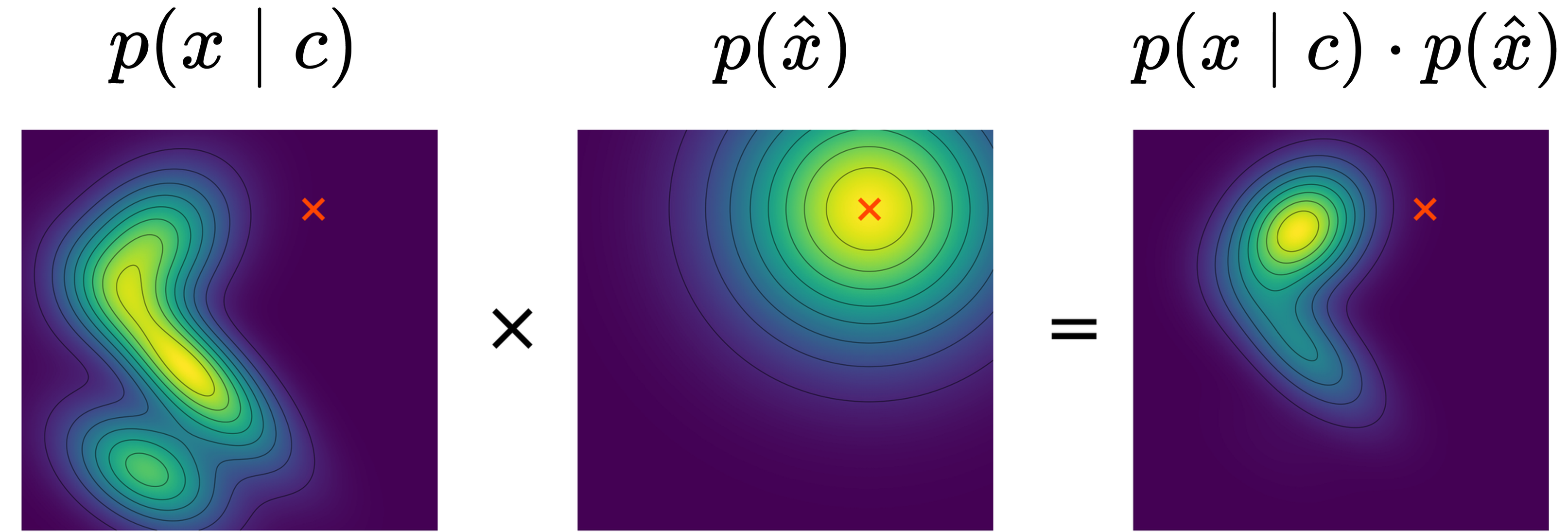}
    \caption{\textbf{Intuition for the Product-of-Experts Regularizer.} The left distribution $p(x \mid c)$ represents the "validity expert," which pushes the latent toward the target counterfactual concepts. The center distribution $p(\hat{x})$ represents the "proximity expert," which anchors the generation to the original input manifold to prevent catastrophic degradation. The resulting product (right) defines a sharpened search space that satisfies both constraints simultaneously.}
    \label{fig:product_of_experts}
\end{figure}

\begin{equation}
    \nabla_x \log\left( q^{(w)}(x_{t-1} \mid x_t, c) \cdot q(x_t \mid \hat{x}) \right) = s^{(w)}(\mathbf{x}_t, t, c) - \frac{(x_t - \sqrt{\bar{\alpha}_{t}}\hat{x})}{(1-\bar{\alpha}_t)}.
\end{equation}
In noise-prediction terms, this yields a guided noise estimate composed of a validity term and a proximity term:
\begin{equation}
    \epsilon_\theta(\mathbf{x}_t, t, c, \hat{x}) = \underbrace{\epsilon^{(w)}_\theta(\mathbf{x}_t, t, c)}_{\text{validity term}} + \underbrace{\frac{(x_t - \sqrt{\bar{\alpha}_{t}}\hat{x})}{\sqrt{1-\bar{\alpha}_t}}}_{\text{proximity term}}.
\end{equation}

\paragraph{Intuition.}
The intuition behind this approach is to provide a "soft anchor" that prevents the diffusion process from wandering into unrealistic regions of the latent space. The validity expert acts as a directional force, while the proximity expert acts as a restorative force, ensuring that only the minimal necessary semantic changes occur. This formulation allows us to balance concept alignment with similarity to the source image without requiring expensive clean-image approximations or multiple diffusion passes.



A practical issue arises in the reverse diffusion process as $t \to 0$: since $\sqrt{1-\bar{\alpha}_t} \to 0$, the proximity term can dominate the sampling trajectory, potentially forcing the reconstruction to collapse prematurely into the original latent $\hat{x}$. To prevent this, we ensure that the anchoring behavior applies only to semantic regions that are irrelevant for changing the model's prediction. 

We achieve this by computing a dynamic, gradient-based mask on-the-fly during each step of the reverse diffusion. First, we derive a denoised estimate of the latent, $z_t$, at timestep $t$:

\begin{equation}
    z_t = \frac{x_t}{\sqrt{\bar{\alpha}_t}} - \frac{\sqrt{1-\bar{\alpha}_t}\,\epsilon_\theta(x_t,t,\emptyset)}{\sqrt{\bar{\alpha}_t}}
\end{equation}

Next, we calculate the gradient of the classifier's prediction with respect to this estimate to identify the features most influential to the target concept. We normalize this gradient to create the saliency map $M_{\text{pre}}$:

\begin{equation}
    M_{\text{pre}} = \frac{\nabla_{z_t} p(c \mid z_t)}{\left\lVert \nabla_{z_t} p(c \mid z_t) \right\rVert_\infty}
\end{equation}

Then, we binarize this map with a threshold $\kappa$ to produce the final spatial mask $M$:

\begin{equation}
    M_{i,j} = 
    \begin{cases} 
      1 & \text{if } M_{\text{pre},i,j} > \kappa \\
      0 & \text{otherwise}
    \end{cases}
\end{equation}

This mask allows the "validity expert" to operate freely in relevant semantic regions (where $M=1$), while the "proximity expert" ensures that the rest of the image (where $M=0$) remains identical to the source. By integrating this calculation directly into the reverse sampling loop, C-VCE enables precisely localized counterfactuals in a single pass.

Finally, we combine both terms using the binary mask $M$ to obtain the final noise prediction $\epsilon_\theta$:
\begin{equation}
    \epsilon_\theta(\mathbf{x}_t, t, c, \hat{x}) = M \odot \epsilon^{(w)}_\theta(\mathbf{x}_t, t, c) + (1-M) \odot \frac{(x_t - \sqrt{\bar{\alpha}_{t}}\hat{x})}{\sqrt{1-\bar{\alpha}_t}}.
\end{equation}

The resulting sampling algorithm is detailed in Algorithm~\ref{alg:sampling}.

\begin{algorithm}
\caption{C-VCE Counterfactual Sampling}
\label{alg:sampling}
\begin{algorithmic}[1]
\State \textbf{Input:} Original image $\hat{x}$, target concepts $c'$, guidance weights $w$, starting timestep $\tau$, mask threshold $\kappa$
\State $\mathbf{x}_{\tau} \sim \mathcal{N}(\mathbf{x}_{\bar{t}}; \sqrt{\bar{\alpha}_{\tau}} \hat{x}, (1 - \bar{\alpha}_{\tau}) \mathbf{I})$
\For{$t = \tau, \ldots, 1$}
    \State $\boldsymbol{\eta} \sim \mathcal{N}(0, \mathbf{I})$ if $t > 1$, else $\boldsymbol{\eta} = 0$
    \State $\boldsymbol{\hat{\epsilon}}^{(w)} = (1+w)\boldsymbol{\epsilon}_\theta(\mathbf{x}_{t},t,c') - w\boldsymbol{\epsilon}_\theta(\mathbf{x}_{t},t,\emptyset)$
    \State $z_t = \frac{\mathbf{x}_t}{\sqrt{\bar{\alpha}_t}} - \frac{\sqrt{1-\bar{\alpha}_t}\,\epsilon_\theta(\mathbf{x}_t,t,\emptyset)}{\sqrt{\bar{\alpha}_t}}$
    \State $M_{\text{pre}} = \frac{\nabla_{z_t} p(c \mid z_t)}{\left\lVert \nabla_{z_t} p(c \mid z_t) \right\rVert_\infty}$
    \State $M_{i,j} \gets \mathbb{I}\left[M_{\text{pre},i,j} > \kappa \right]$
    \State $\boldsymbol{\hat{\epsilon}} = M \odot \boldsymbol{\hat{\epsilon}}^{(w)} + (1-M) \odot   \frac{(\mathbf{x}_t - \sqrt{\bar{\alpha}_{t}}\hat{x})}{\sqrt{1-\bar{\alpha}_t}} $
    \State $\mathbf{x}_{t-1} = \frac{1}{\sqrt{\alpha_t}} \left( \mathbf{x}_t - \frac{1 - \alpha_t}{\sqrt{1 - \bar{\alpha}_t}}\boldsymbol{\hat{\epsilon}} \right) + \sigma_t \boldsymbol{\eta}$
\EndFor
\State \Return $\mathbf{x}_0$
\end{algorithmic}
\end{algorithm}



\subsection{Joint Training Objective}

Before describing the counterfactual sampling process, we define the training procedure for the C-VCE framework. Unlike standard DDPMs that focus solely on noise reconstruction, C-VCE is trained using a joint optimization strategy that simultaneously learns the data manifold and the semantic concept space. We train the model by minimizing a composite loss function $\mathcal{L}_{\text{total}}$, which balances high-fidelity image synthesis with accurate concept inference:

\begin{equation}
    \mathcal{L}_{\text{total}} = \lambda_{\text{mse}} \mathcal{L}_{\text{noise}} + \lambda_{\text{cbm}} \mathcal{L}_{\text{concept}}
\end{equation}

The first term, $\mathcal{L}_{\text{noise}}$, is the conditional epsilon-prediction loss. During training, the model is conditioned on the ground-truth concept vector $c$ extracted from the dataset, ensuring the denoising process is semantically aligned from the outset:

\begin{equation}
    \mathcal{L}_{\text{noise}} = \mathbb{E}_{x_0, \epsilon \sim \mathcal{N}(0, \mathbf{I}), t} \left[ \left\| \epsilon - \epsilon_\theta(x_t, t, c) \right\|^2_2 \right]
\end{equation}

The second term, $\mathcal{L}_{\text{concept}}$, is the binary cross-entropy loss for the concept bottleneck, which ensures the internal classifier $f_c$ accurately predicts the attributes $c$:

\begin{equation}
    \mathcal{L}_{\text{concept}} = - \mathbb{E} \left[ \sum_{i=1}^k c_i \log(\hat{c}_i) + (1-c_i) \log(1-\hat{c}_i) \right]
\end{equation}

where $\hat{c} = \sigma(f_c(h_{\text{pre}}))$ is the predicted concept vector. This joint optimization ensures that the bottleneck representation is optimized for both reconstruction and semantic grounding. By learning this internal concept mapping during training, the framework provides a reliable foundation for the counterfactual interventions performed during the sampling stage, where the model can then be steered using modified concept targets.

\section{Experiments}
\label{sec:experiments}

In this section, we present evaluation results of the proposed C-VCE framework. We begin by establishing general performance benchmarks through quantitative metrics and visual fidelity comparisons on the CelebA dataset \cite{CelebA:liu2015faceattributes}. Next, we analyze the stability of the model by examining the Pareto efficiency and trade-offs between editability and image quality. Finally, we investigate the semantic reasoning capabilities of our method, specifically focusing on its behavior when encountering highly correlated or conflicting attributes.

\subsection{Experimental Setup}

To foster a reproducible evaluation, our framework utilizes the pre-trained Stable Diffusion VAE \cite{stable:esser2024scalingrectifiedflowtransformers} for all latent space encoding and decoding tasks. The training process is governed by the AdamW optimizer \cite{adamw:loshchilov2019decoupledweightdecayregularization} with a learning rate of $6 \times 10^{-5}$, coupled with a cosine annealing scheduler \cite{cosine:loshchilov2017sgdrstochasticgradientdescent} featuring 500 warmup steps. We employ a batch size of 64 during the training phase and 72 for evaluation on the CelebA dataset. Training is conducted over 300 epochs, utilizing fp16 automatic mixed precision (AMP) to optimize computational efficiency. To maintain consistency across all experimental runs, a fixed random seed of 1265.

Our hyperparameter selection is the result of an extensive grid search designed to navigate the validity-proximity trade-off. For the proposed C-VCE method, we evaluated guidance weights $w \in \{1, 2, 3, 4, 5\}$ while fixing the noise timestep $\tau=200$, a mask threshold $\kappa=0.1$, and an intervention strength $\alpha=1.0$. Based on the resulting Pareto frontier analysis, we identified $w=3$ as the primary configuration for final comparisons due to its optimal balance of validity and proximity. For the L-DVCE baseline, we tested intervention strengths $C_c \in \{0.03, 0.04, 0.05, 0.07, 0.1\}$ with a fixed distance strength $C_d=0.15$ and noise timestep $\tau=200$. We selected $C_c=0.04$ as the best-performing baseline configuration to ensure a fair comparison where both models operate at high success rates.

\subsection{General Results and Visual Fidelity}

We evaluate the performance of C-VCE against the competitive L-DVCE baseline using a multi-faceted metric suite to capture the nuances of counterfactual generation.

\paragraph{Metrics and Evaluation Protocol.}
To provide a holistic assessment, we compute the following metrics:
\begin{itemize}
    \item \textbf{Closeness ($l_1, l_2, l_{1.5}$):} Measures pixel-level distance between input and edited images to assess identity preservation.
    \item \textbf{Realism (sFID):} Uses sliding-window FID  to assess local perceptual quality. Standard FID often misses artifacts in counterfactuals due to their high similarity to the original. sFID addresses this by splitting samples into subsets and comparing local windows to ensure the edits match the target distribution's spatial coherence.
    \item \textbf{Validity (Flip Rate):} Employs an external classifier to measure the success rate of inducing the target attribute.
    \item \textbf{Reliability (Failure Rate):} A robustness metric measuring the percentage of samples where an off-the-shelf face detector (e.g., DeepFace \cite{DEEP:FACE:serengil2024lightface} ) fails, indicating catastrophic image degradation.
    \item \textbf{Identity (ID \%)}: Measures the preservation of the subject's identity by comparing face embeddings of the original and edited images using a pre-trained VGG-Face network~\cite{DEEP:FACE:serengil2024lightface}. A higher percentage indicates that the person's unique facial features remain recognizable after the counterfactual intervention.
\end{itemize}

\paragraph{Quantitative Comparison.} Table \ref{tab:main_results} presents the consolidated results across all primary metrics. C-VCE consistently outperforms the baseline in all proximity metrics. Notably, it achieves a 31\% reduction in $l_1$ error (0.0156 vs 0.0227). While the Baseline maintains a marginally higher Flip Rate (0.9984) and Identity preservation (95.1\% via VGG-Face), it suffers from a significantly higher Failure Rate (3.84\%). C-VCE reduces this to 2.44\% (a 36.4\% relative improvement) and achieves superior Realism (11.42 sFID), indicating better perceptual quality and manifold consistency on challenging samples.

\begin{table}[h!]
\caption{Consolidated results. Standard deviations are shown in parentheses. C-VCE outperforms the baseline in closeness metrics and Failure Rate (Fail), balancing high validity with superior photorealism.}
\label{tab:main_results}
\centering
\small
\resizebox{\textwidth}{!}{%
\begin{tabular}{|l|ccc|c|cc|c|}
\cline{2-8}
\multicolumn{1}{c|}{} & \multicolumn{3}{c|}{\textbf{Closeness}} & \textbf{Validity} & \multicolumn{2}{c|}{\textbf{Identity}} & \textbf{Realism} \\
\hline
\textbf{Method} & $l_1 \downarrow$ & $l_{1.5} \downarrow$ & $l_2 \downarrow$ & FR $\uparrow$ & ID \% $\uparrow$ & Fail \% $\downarrow$ & sFID $\downarrow$ \\
\hline
\hline
L-DVCE $_{CelebA}$ & 0.0227 \scriptsize{(0.004)}& 0.0319 \scriptsize{(0.005)}& 0.0428 \scriptsize{(0.007)}& \textbf{0.9984} & \textbf{95.1} & 3.84 & 11.65 \\
C-VCE $_{CelebA}$ & \textbf{0.0156} \scriptsize{(0.003)} & \textbf{0.0267} \scriptsize{(0.006)} & \textbf{0.0389} \scriptsize{(0.01)} & 0.9788 & 94.5 & \textbf{2.44} & \textbf{11.42} \\
\hline
\end{tabular}%
}
\end{table}

\begin{figure}[h!]
    \centering
    \includegraphics[width=1\linewidth]{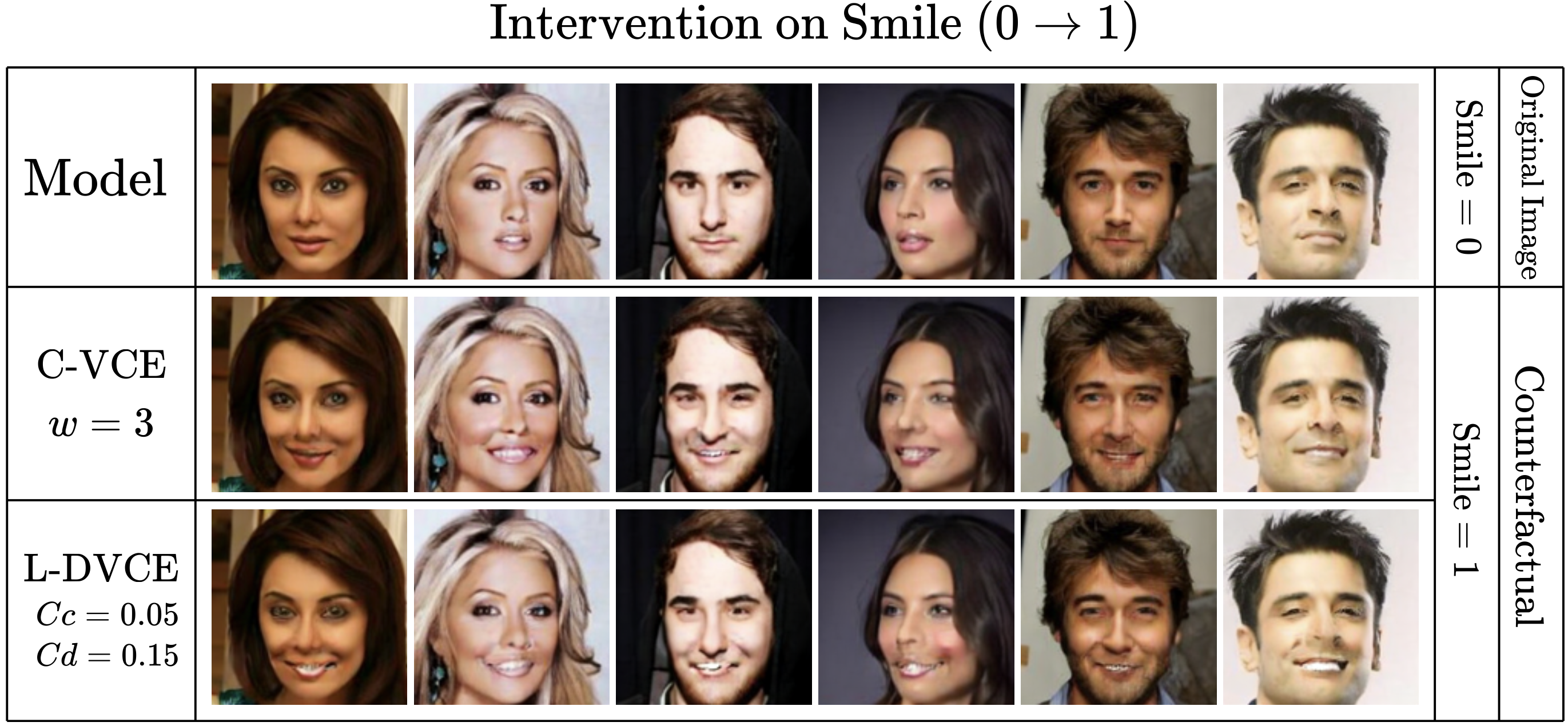}
    \caption{\textbf{Visual Fidelity Comparison.} From top to bottom: (a) Original Input, (b) C-VCE ($w=3$), and (c) L-DVCE baseline. The Baseline exhibits catastrophic artifacts such as glowing teeth, whereas C-VCE maintains photorealism.}
    \label{fig:smile_qualitative}
\end{figure}

\paragraph{Qualitative Analysis.}
Visual inspection confirms the metric improvements. Figure \ref{fig:smile_qualitative} compares the models on the 'Smiling' attribute. The L-DVCE baseline frequently introduces high-frequency spectral artifacts, most notably "glowing" or oversaturated teeth and plastic-like skin deformations that do not respect the original facial geometry. These artifacts are particularly visible in the fourth and sixth columns of Figure \ref{fig:smile_qualitative}(c). 

In contrast, C-VCE synthesizes more naturalistic expressions that maintain the subject's original lighting and identity. The results in row (b) demonstrate that C-VCE avoids the saturation issues seen in the baseline, corroborating the lower sFID scores and higher reliability metrics.

\subsection{Trade-off Analysis: Stability vs. Plasticity}

To understand the behavior of the models beyond average scores, we analyze the Pareto frontiers (efficiency) and latent space stability.

\begin{figure}[h!]
    \centering
    \begin{subfigure}[b]{0.48\linewidth}
        \centering
        \includegraphics[width=\linewidth]{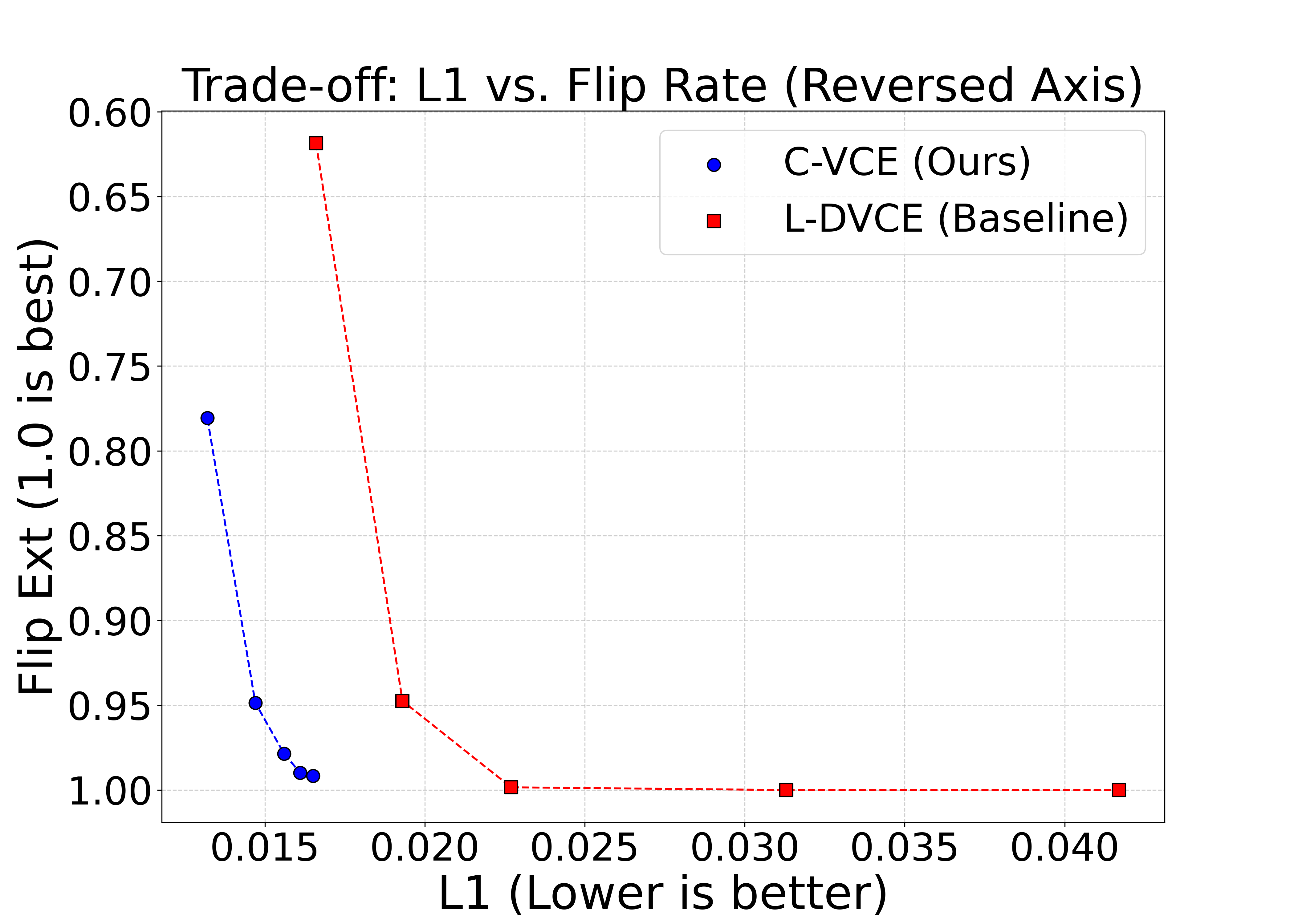}
        \caption{$l_1$ vs. Flip Rate}
    \end{subfigure}
    \hfill
    \begin{subfigure}[b]{0.48\linewidth}
        \centering
        \includegraphics[width=\linewidth]{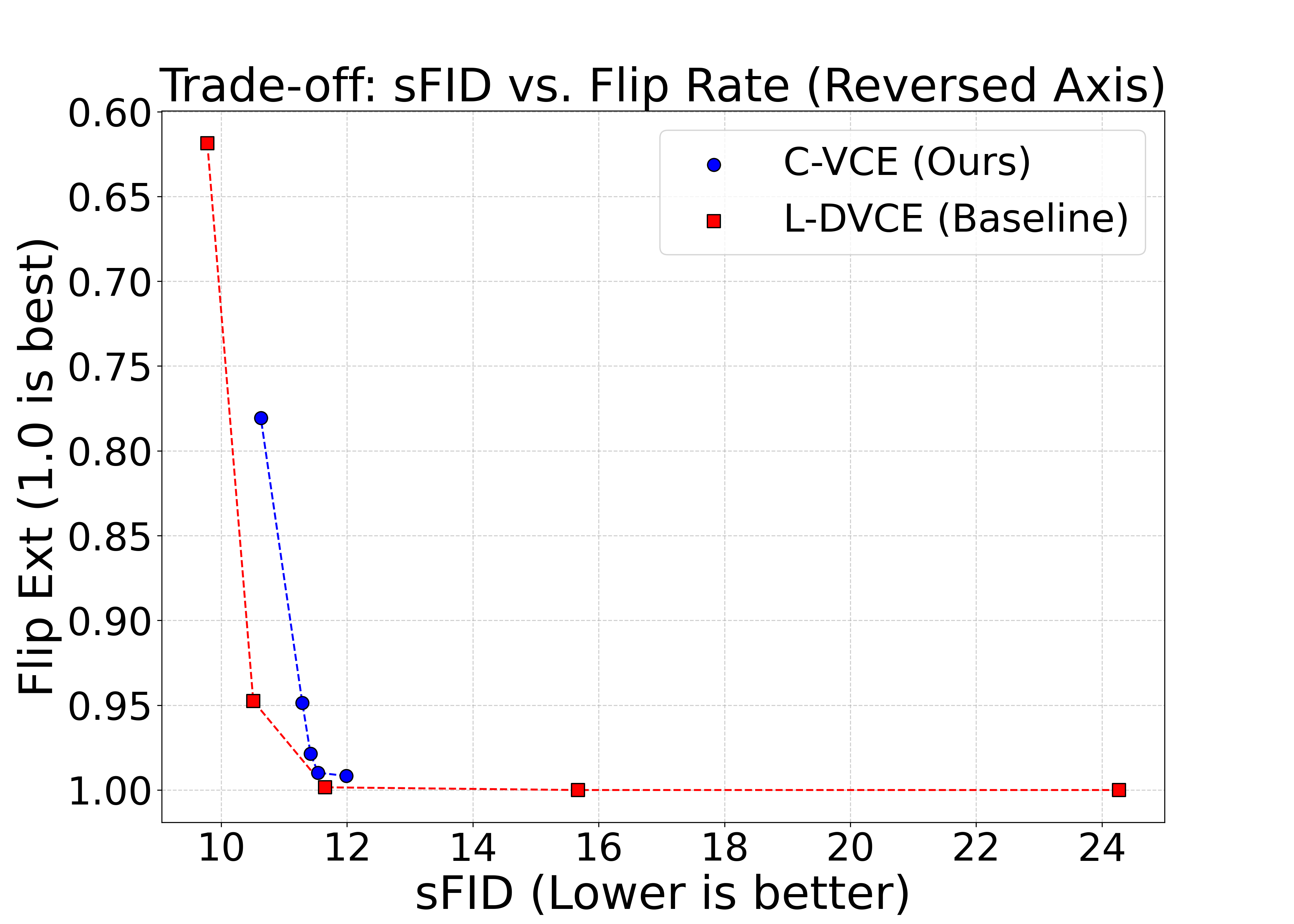}
        \caption{sFID vs. Flip Rate}
    \end{subfigure}
    
    \vspace{0.2cm}
    
    \begin{subfigure}[b]{0.48\linewidth}
        \centering
        \includegraphics[width=\linewidth]{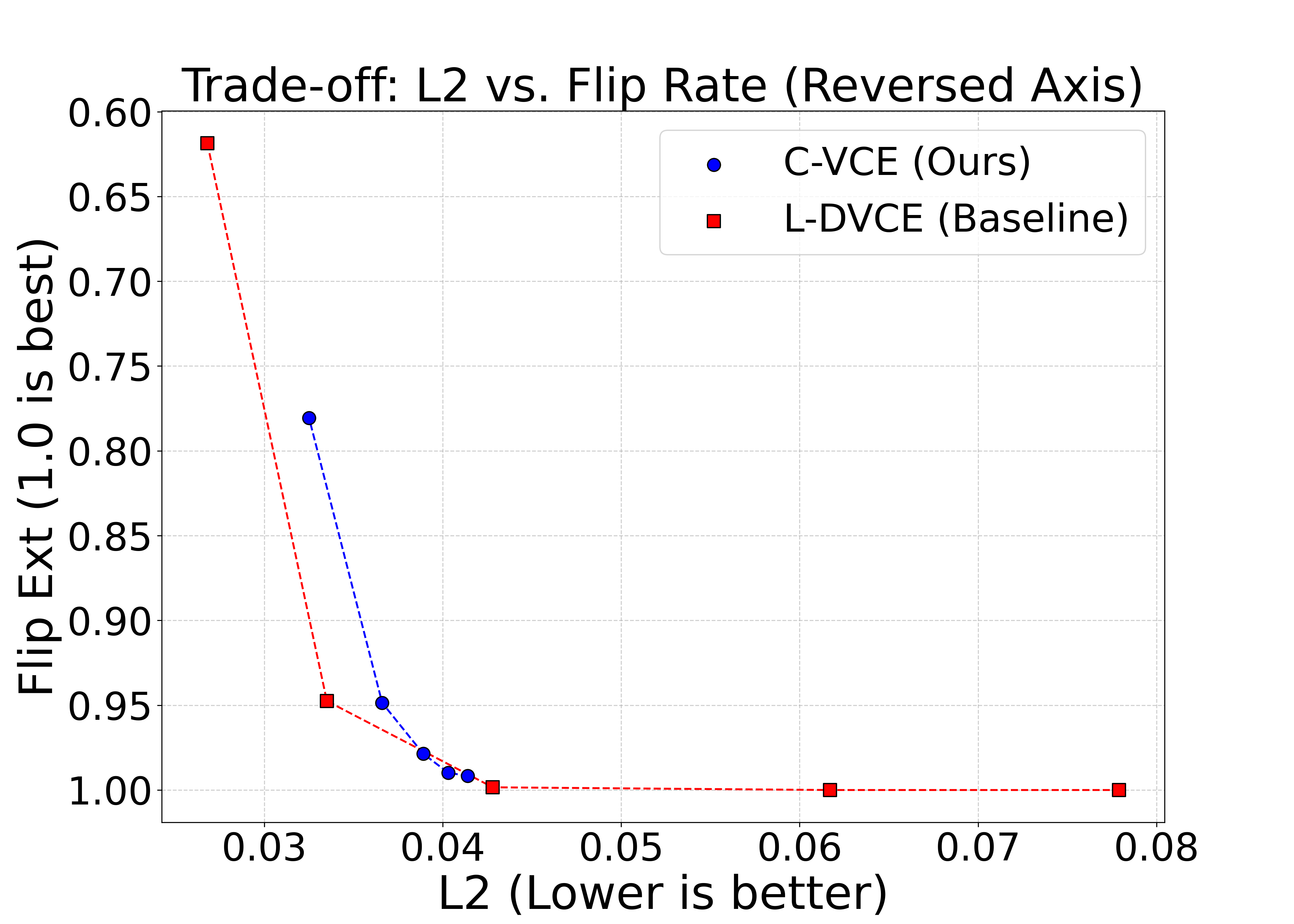}
        \caption{$l_2$ vs. Flip Rate}
    \end{subfigure}
    \hfill
    \begin{subfigure}[b]{0.48\linewidth}
        \centering
        \includegraphics[width=\linewidth]{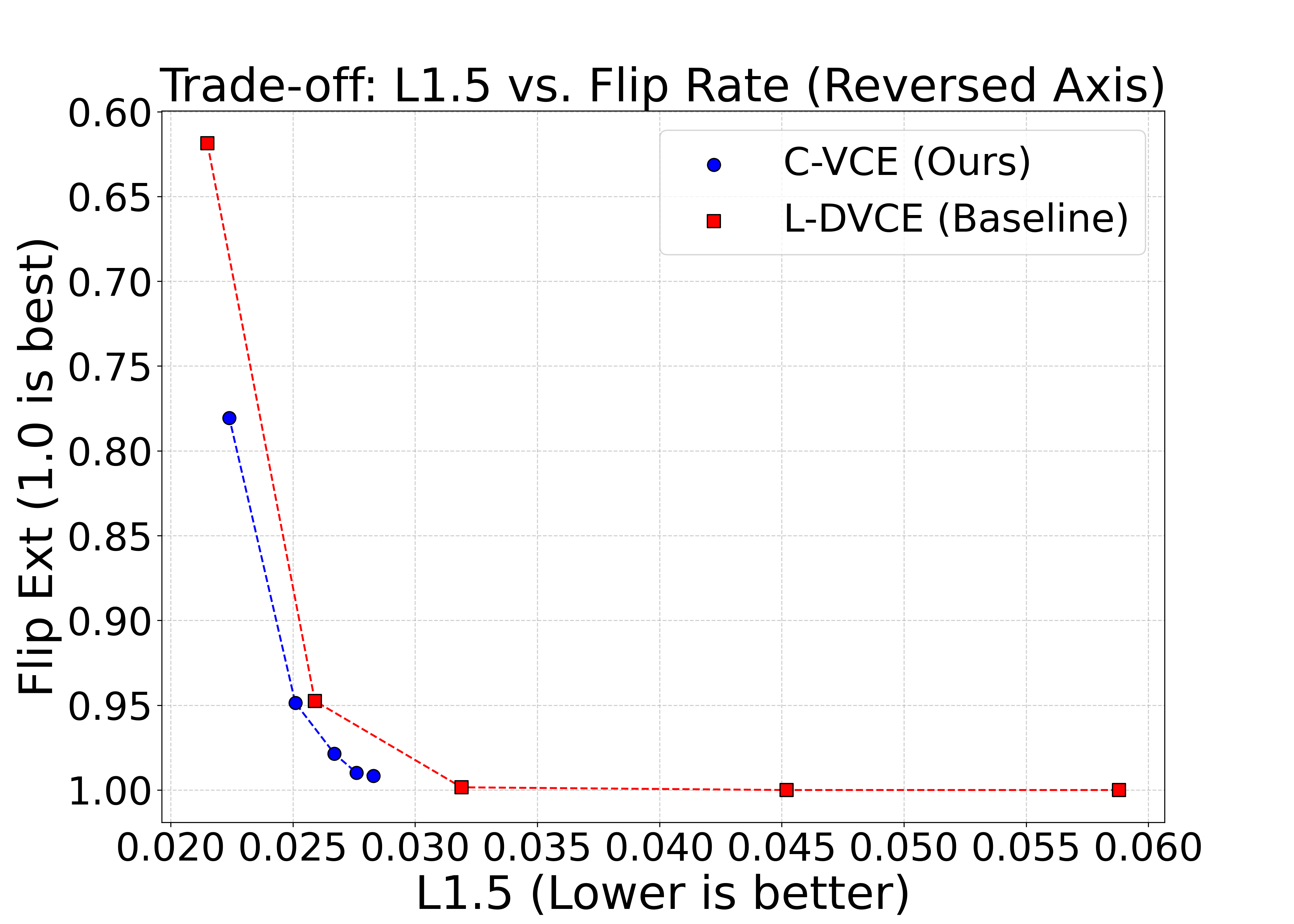}
        \caption{$l_{1.5}$ vs. Flip Rate}
    \end{subfigure}

    \caption{\textbf{Pareto Efficiency.} C-VCE (Blue) maintains high quality (low sFID) even at max intensity, whereas the Baseline (Red) degrades severely.}
    \label{fig:quantitative_tradeoffs}
\end{figure}

\paragraph{Pareto Frontiers.}
In Figure \ref{fig:quantitative_tradeoffs}, we plot the Editability (Flip Rate) against Error metrics. 
The plots reveal that C-VCE (Blue) consistently dominates the frontier, achieving lower reconstruction error ($l_1, l_{1.5}$) than the Baseline (Red) for any given success rate. Crucially, Figure \ref{fig:quantitative_tradeoffs}(b) exposes a critical instability in the Baseline, i.e., as it pushes for perfect Flip Rates (1.0), its sFID explodes to $>24$, indicating mode collapse.

\paragraph{Latent Stability.}
We further investigate model stability by visualizing the latent traversal as a function of increasing intervention weights. In this context, each generated counterfactual represents a specific operating point on the Pareto frontier (see Fig.~\ref{fig:quantitative_tradeoffs}), where the model must navigate the competing objectives of attribute validity and image proximity.

\begin{figure}[h!]
    \centering
    \begin{subfigure}[b]{\linewidth}
        \centering
        \includegraphics[width=\linewidth]{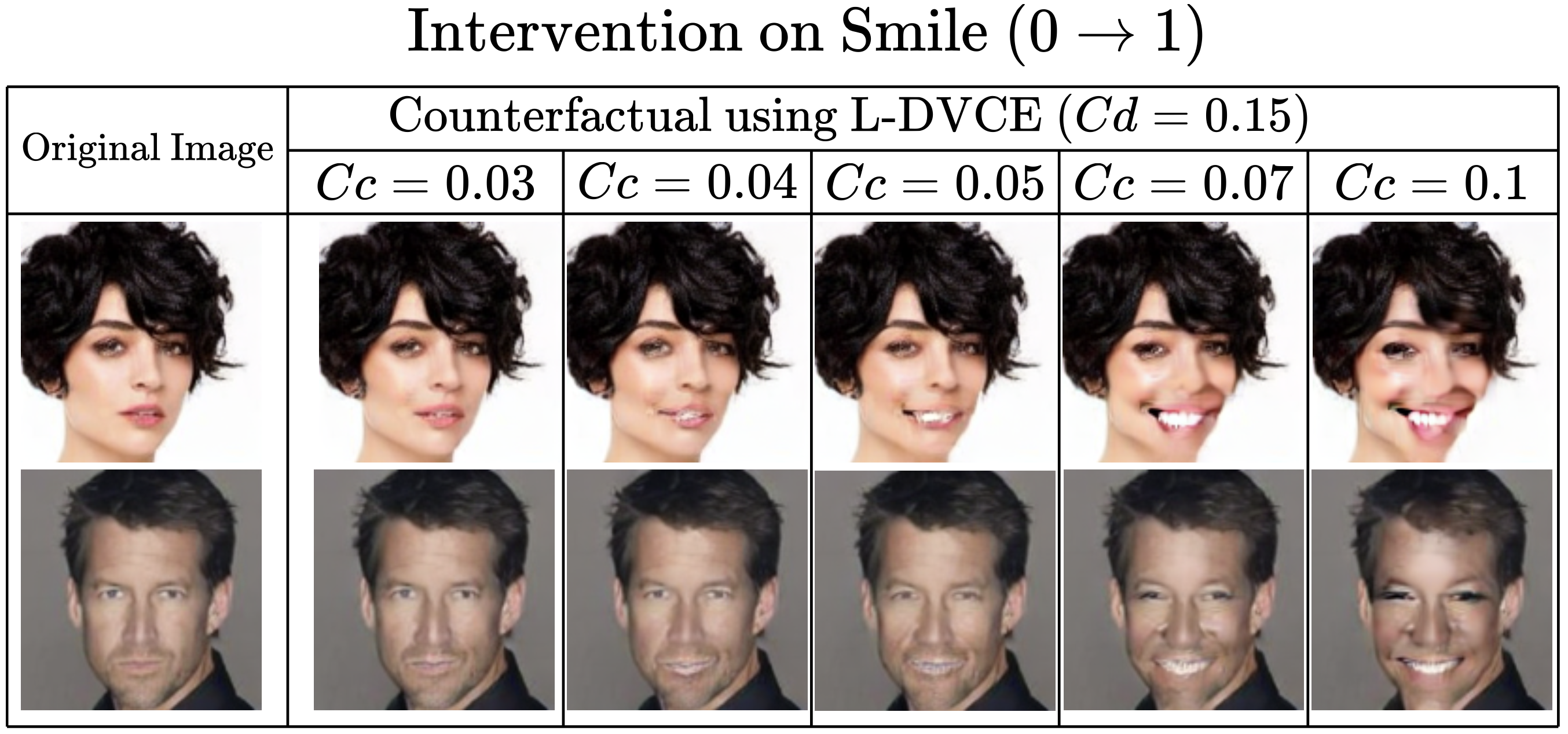} 
        \caption{Baseline (L-DVCE) Instability: Increasing $C_c$ leads to manifold breakdown.}
        \label{fig:dvce_tradeoff}
    \end{subfigure}
    
    \vspace{0.4cm}
    
    \begin{subfigure}[b]{\linewidth}
        \centering
        \includegraphics[width=\linewidth]{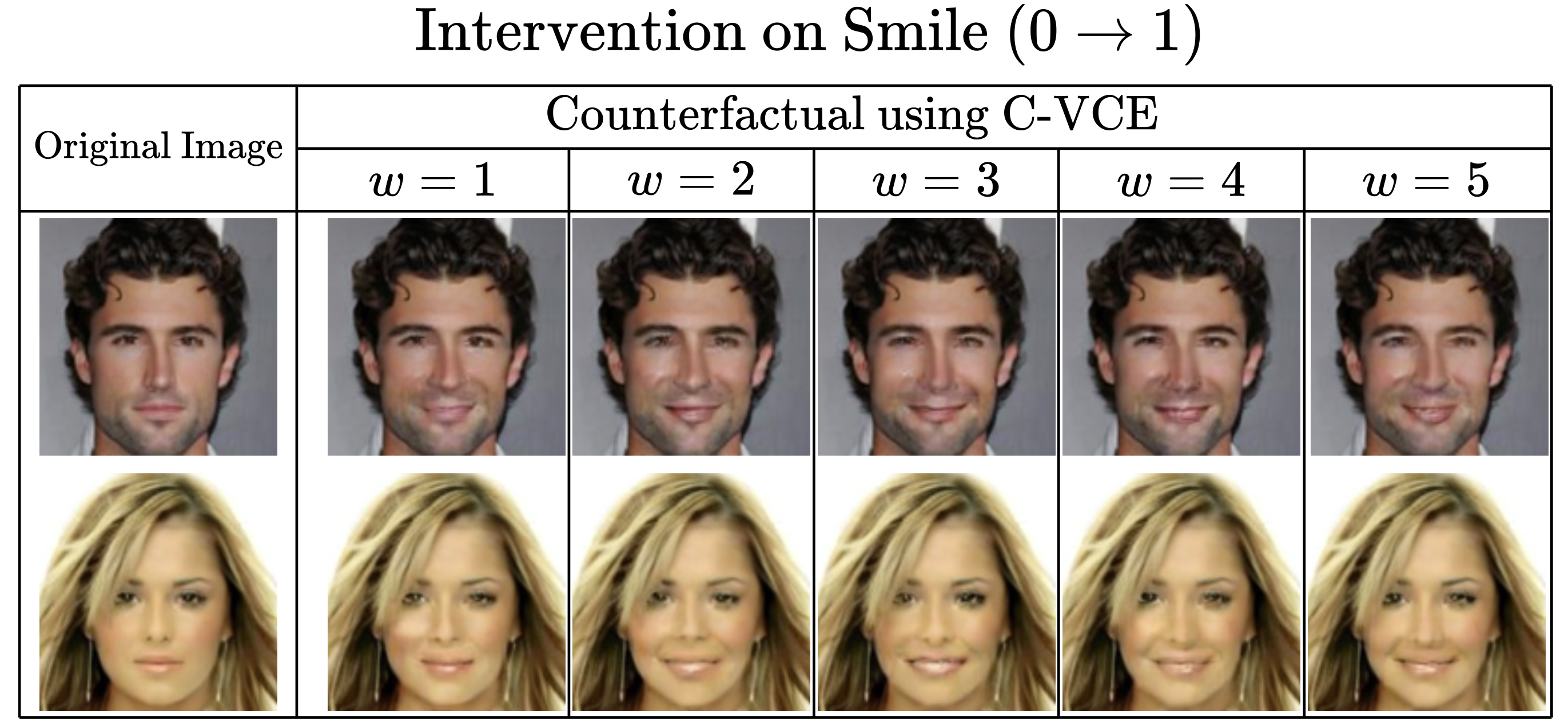} 
        \caption{Ours (C-VCE) Stability: Increasing $w$ yields smooth, realistic transitions.}
        \label{fig:cbm_tradeoff}
    \end{subfigure}
    \caption{\textbf{Latent Traversal Analysis.} The Baseline (a) suffers from saturation artifacts at high intensity, while C-VCE (b) produces artifact-free transitions by preserving the image manifold.}
    \label{fig:latent_traversal}
\end{figure}

As illustrated in Figure \ref{fig:latent_traversal}a, the L-DVCE baseline exhibits severe "saturation artifacts" as the intervention strength $C_c$ scales from $0.03$ to $0.1$. At high intensities, the optimization effectively "breaks" the natural image manifold to satisfy the classifier, resulting in unnatural white patches and a loss of identity preservation. Conversely, C-VCE (Fig. \ref{fig:latent_traversal}b) demonstrates smooth, continuous interpolation as the guidance weight $w$ increases from $1$ to $5$. Our method maintains structural coherence even at maximum intensity, corroborating the lower sFID and $l_1$ scores observed in the quantitative trade-off graphs.

\subsection{Semantic Consistency and Attribute Correlation}

An important aspect of counterfactual generation is how a model handles the inherent statistical biases of its training data. We investigate this by examining a scenario involving highly conflicting attributes: adding a 'Beard' to 'Female' subjects—a combination that is statistically rare in the CelebA dataset. This experiment reveals whether the generation process is constrained by the learned attribute correlations or if it forces the edit regardless of the dataset's underlying distribution.

\paragraph{Analysis of Conflicting Attributes.}

\begin{figure}[h!]
    \centering
    \includegraphics[width=1\linewidth]{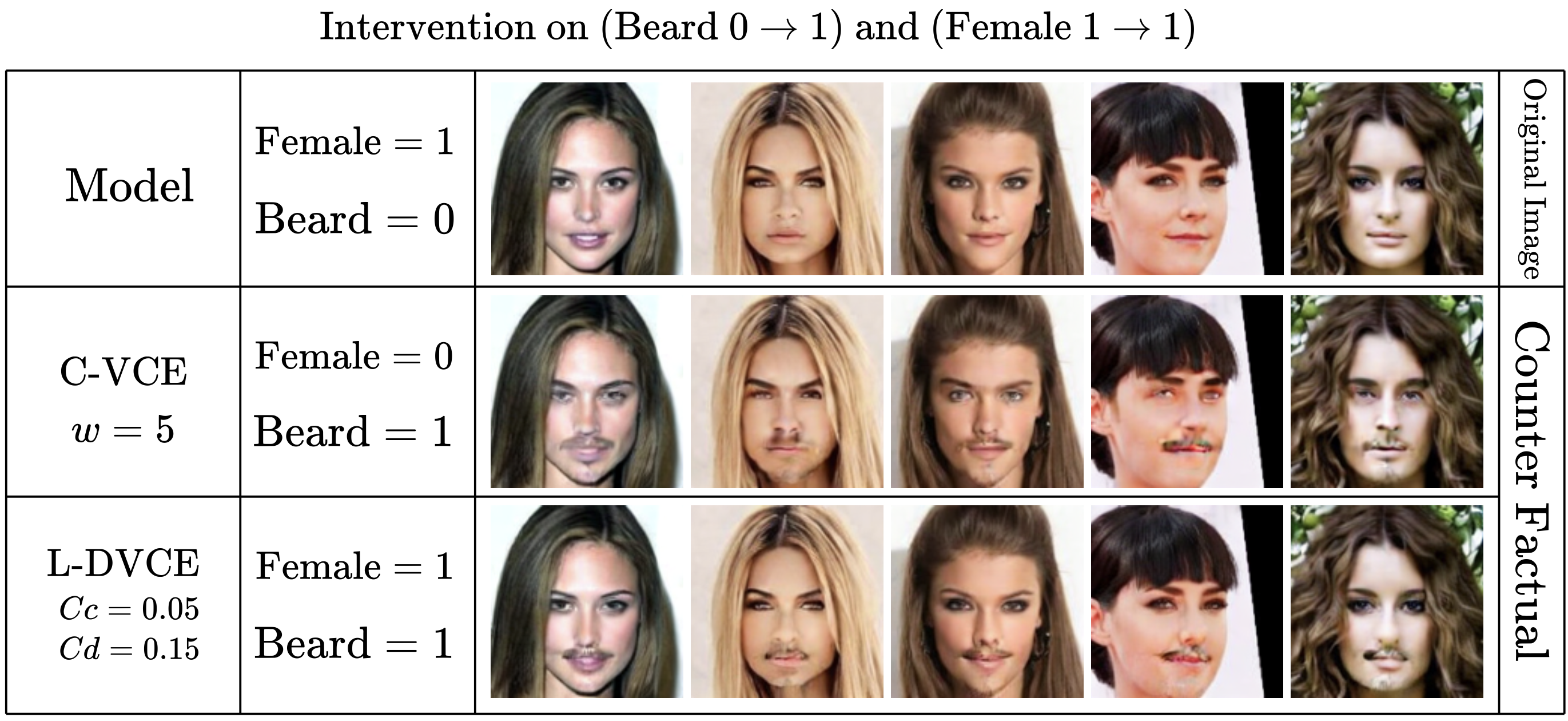}
    \caption{\textbf{Counterfactual Intervention (Beard on Female).} The Baseline (bottom) creates semantically inconsistent "Bearded Females" to satisfy the classifier. C-VCE (middle) preserves the subject's identity and gender identity by refusing the intervention, thereby respecting the natural data manifold.}
    \label{fig:correlation_matrix}
\end{figure}

As shown in Table \ref{tab:correlation_failure}, the L-DVCE baseline exhibits a high "Flip Rate" (0.7483). While this indicates the classifier's condition is met, visual inspection in Figure \ref{fig:correlation_matrix} shows that the baseline model pastes facial hair onto female subjects, resulting in artifacts that reduce visual plausibility.


In contrast, C-VCE yields a near-zero Flip Rate (0.0127). This behavior reflects the statistical bias of the training data. Because C-VCE captures the joint distribution of attributes in CelebA, it restricts the generation of combinations that are statistically rare in the dataset (such as adding a beard to a female subject). Consequently, the model suppresses the edit and preserves the original image features rather than generating an image that contradicts the dataset's learned attribute correlations.

\begin{table}[h!]
\caption{\textbf{Semantic Consistency Results.} The baseline forces the edit (High FR) despite the statistical impossibility, whereas C-VCE preserves the manifold by refusing semantically inconsistent edits.}
\centering
\small
\begin{tabular}{|l|ccc|c|}
\cline{2-5}
\multicolumn{1}{c|}{} & \multicolumn{3}{c|}{\textbf{Pixel Preservation}} & \multicolumn{1}{c|}{\textbf{Semantic Behavior}} \\
\hline
\textbf{Method} & $l_1 \downarrow$ & $l_{1.5} \downarrow$ & $l_2 \downarrow$ & \textbf{Flip Rate} \\
\hline
\hline
L-DVCE (Baseline) & 0.0205 & \textbf{0.0271} & \textbf{0.0347} & 0.7483 \textit{(Forced Edit)} \\
C-VCE (Ours)    & \textbf{0.0182} & 0.0295 & 0.0423 & \textbf{0.0127} \textit{(Edit Suppression)} \\
\hline
\end{tabular}
\label{tab:correlation_failure}
\end{table}

It is important to note that within the CelebA dataset, long hair is not exclusively correlated with the female attribute, as the distribution contains samples of males with long hair. Consequently, when performing an intervention targeting the male attribute, C-VCE maintains the subject's hair length while adjusting features that have a stronger statistical correlation with the target attribute in the dataset (such as the removal of makeup). This demonstrates how C-VCE's generated edits are strictly driven by the statistical biases and attribute correlations present in the training data.

\section{Limitations and Future Work}
\label{sec:discussion}

While C-VCE demonstrates superior alignment with the training data distribution and reliability compared to traditional latent-optimization methods, it possesses specific limitations that offer avenues for future research. A primary technical constraint is the computational latency inherent to the DDPM framework. Because the generation process requires a full reverse diffusion sequence over $T$ timesteps, counterfactual synthesis is significantly slower than single-pass generative models. Future work will focus on optimizing this sampling process, investigating the integration of our guidance mechanism into accelerated sampling frameworks or Latent Consistency Models (LCMs) to reduce inference time while maintaining high fidelity.

Additionally, our current implementation utilizes a fixed guidance weight $w$ across all samples. While our grid search identified $w=3$ as a robust "sweet spot" for the CelebA dataset, we observe that the optimal intervention intensity can be image-dependent. We propose exploring adaptive weighting strategies where $w$ is dynamically adjusted based on the initial distance of a sample from the target decision boundary. 

Furthermore, we aim to more deeply exploit the underlying CBM architecture. Since the concept embeddings contain rich, high-dimensional information about the image's inherent features, they represent a latent goldmine for extracting more granular, localized concepts during the diffusion process. Finally, while C-VCE has shown significant promise on face-centric data (CelebA), further investigation is required to evaluate its performance on more heterogeneous and complex datasets to ensure the "Semantic Refusal" mechanism scales to non-facial domains.

\section{Conclusion}

In this work, we presented C-VCE, a framework for counterfactual generation that utilizes Concept Bottleneck Models within a diffusion-based architecture to prioritize alignment with the training data distribution. Our evaluation on the CelebA  demonstrates that while competitive baselines may achieve marginally higher success rates, they often do so at the cost of catastrophic image degradation and high-frequency artifacts. By navigating the validity-proximity trade-off across multiple distance norms ($l_1, l_{1.5}, l_2$), C-VCE ensures that generated counterfactuals remain plausible and interpretable. This inherent respect for the learned data distribution makes C-VCE a promising candidate for high-stakes visual domains, such as medical diagnostics, where biological and structural integrity are paramount.

\begin{credits}
\subsubsection{\ackname} 
The study was funded by the proejct EXperiMental, within the framework of the Second Swiss Contribution Multilateral Academic
Projects (MAPS) by the SNSF (GA no. 230189). M. Gjoreski is funded by SNSF through the project XAI-PAC (GA no. 216405). D. Kirilenko is funded by the project TRUST-ME (GA no. 214991).

\subsubsection{\discintname}
The authors have no competing interests to declare.
\end{credits}

\bibliographystyle{splncs04}
\bibliography{references}

@article{exp:healthcare:saraswat2022explainable,
  title={Explainable AI for healthcare 5.0: opportunities and challenges},
  author={Saraswat, Deepti and Bhattacharya, Pronaya and Verma, Ashwin and Prasad, Vivek Kumar and Tanwar, Sudeep and Sharma, Gulshan and Bokoro, Pitshou N and Sharma, Ravi},
  journal={IEEe Access},
  volume={10},
  pages={84486--84517},
  year={2022},
  publisher={IEEE}
}

@article{exp:car:kuznietsov2024explainable,
  title={Explainable AI for safe and trustworthy autonomous driving: A systematic review},
  author={Kuznietsov, Anton and Gyevnar, Balint and Wang, Cheng and Peters, Steven and Albrecht, Stefano V},
  journal={IEEE Transactions on Intelligent Transportation Systems},
  year={2024},
  publisher={IEEE}
}

@misc{expl:das2020opportunitieschallengesexplainableartificial,
      title={Opportunities and Challenges in Explainable Artificial Intelligence (XAI): A Survey}, 
      author={Arun Das and Paul Rad},
      year={2020},
      eprint={2006.11371},
      archivePrefix={arXiv},
      primaryClass={cs.CV},
      url={https://arxiv.org/abs/2006.11371}, 
}

@article{right:to:explain:Goodman_2017,
   title={European Union Regulations on Algorithmic Decision Making and a “Right to Explanation”},
   volume={38},
   ISSN={2371-9621},
   url={http://dx.doi.org/10.1609/aimag.v38i3.2741},
   DOI={10.1609/aimag.v38i3.2741},
   number={3},
   journal={AI Magazine},
   publisher={Wiley},
   author={Goodman, Bryce and Flaxman, Seth},
   year={2017},
   month=sep, pages={50–57} }

@misc{DDPM:ho2020denoisingdiffusionprobabilisticmodels,
      title={Denoising Diffusion Probabilistic Models}, 
      author={Jonathan Ho and Ajay Jain and Pieter Abbeel},
      year={2020},
      eprint={2006.11239},
      archivePrefix={arXiv},
      primaryClass={cs.LG},
      url={https://arxiv.org/abs/2006.11239}, 
}

@book{counterfactual:molnar2020interpretable,
  title={Interpretable machine learning},
  author={Molnar, Christoph},
  year={2020},
  publisher={Lulu. com}
}

@misc{counter:factual:basics:wachter2018counterfactualexplanationsopeningblack,
      title={Counterfactual Explanations without Opening the Black Box: Automated Decisions and the GDPR}, 
      author={Sandra Wachter and Brent Mittelstadt and Chris Russell},
      year={2018},
      eprint={1711.00399},
      archivePrefix={arXiv},
      primaryClass={cs.AI},
      url={https://arxiv.org/abs/1711.00399}, 
}

@article{counter:factual:DBLP:journals/datamine/Guidotti24,
  author={Riccardo Guidotti},
  title={Counterfactual explanations and how to find them: literature review and benchmarking},
  year={2024},
  month={September},
  cdate={1725148800000},
  journal={Data Min. Knowl. Discov.},
  volume={38},
  number={5},
  pages={2770-2824},
  url={https://doi.org/10.1007/s10618-022-00831-6}
}

@misc{curse:szegedy2014intriguingpropertiesneuralnetworks,
      title={Intriguing properties of neural networks}, 
      author={Christian Szegedy and Wojciech Zaremba and Ilya Sutskever and Joan Bruna and Dumitru Erhan and Ian Goodfellow and Rob Fergus},
      year={2014},
      eprint={1312.6199},
      archivePrefix={arXiv},
      primaryClass={cs.CV},
      url={https://arxiv.org/abs/1312.6199}, 
}

@misc{rebust:santurkar2019imagesynthesissinglerobust,
      title={Image Synthesis with a Single (Robust) Classifier}, 
      author={Shibani Santurkar and Dimitris Tsipras and Brandon Tran and Andrew Ilyas and Logan Engstrom and Aleksander Madry},
      year={2019},
      eprint={1906.09453},
      archivePrefix={arXiv},
      primaryClass={cs.CV},
      url={https://arxiv.org/abs/1906.09453}, 
}

@misc{rebust:augustin2020adversarialrobustnessinoutdistribution,
      title={Adversarial Robustness on In- and Out-Distribution Improves Explainability}, 
      author={Maximilian Augustin and Alexander Meinke and Matthias Hein},
      year={2020},
      eprint={2003.09461},
      archivePrefix={arXiv},
      primaryClass={cs.LG},
      url={https://arxiv.org/abs/2003.09461}, 
}

@misc{no:robust:prach2025intriguingpropertiesrobustclassification,
      title={Intriguing Properties of Robust Classification}, 
      author={Bernd Prach and Christoph H. Lampert},
      year={2025},
      eprint={2412.04245},
      archivePrefix={arXiv},
      primaryClass={cs.CV},
      url={https://arxiv.org/abs/2412.04245}, 
}

@misc{adv-example:goodfellow2015explainingharnessingadversarialexamples,
      title={Explaining and Harnessing Adversarial Examples}, 
      author={Ian J. Goodfellow and Jonathon Shlens and Christian Szegedy},
      year={2015},
      eprint={1412.6572},
      archivePrefix={arXiv},
      primaryClass={stat.ML},
      url={https://arxiv.org/abs/1412.6572}, 
}

@article{vae:issue:zhao2017toward,
  title={Towards deeper understanding of variational autoencoding models},
  author={Zhao, Shengjia and Song, Jiaming and Ermon, Stefano},
  journal={arXiv preprint arXiv:1702.08658},
  year={2017}
}

@misc{GAN:original:goodfellow2014generativeadversarialnetworks,
      title={Generative Adversarial Networks}, 
      author={Ian J. Goodfellow and Jean Pouget-Abadie and Mehdi Mirza and Bing Xu and David Warde-Farley and Sherjil Ozair and Aaron Courville and Yoshua Bengio},
      year={2014},
      eprint={1406.2661},
      archivePrefix={arXiv},
      primaryClass={stat.ML},
      url={https://arxiv.org/abs/1406.2661}, 
}

@InProceedings{condGAN:Samangouei_2018_ECCV,
author = {Samangouei, Pouya and Saeedi, Ardavan and Nakagawa, Liam and Silberman, Nathan},
title = {ExplainGAN: Model Explanation via Decision Boundary Crossing Transformations},
booktitle = {Proceedings of the European Conference on Computer Vision (ECCV)},
month = {September},
year = {2018}
}

@article{daniil:kirilenko2024generative,
  title={Generative Models for Counterfactual Explanations},
  author={Kirilenko, Daniil and Barbiero, Pietro and Gjoreski, Martin and Lu{\v{s}}trek, Mitja and Langheinrich, Marc},
  journal={View Article},
  year={2024}
}

@misc{semantic:gan:sauer2021counterfactualgenerativenetworks,
      title={Counterfactual Generative Networks}, 
      author={Axel Sauer and Andreas Geiger},
      year={2021},
      eprint={2101.06046},
      archivePrefix={arXiv},
      primaryClass={cs.LG},
      url={https://arxiv.org/abs/2101.06046}, 
}

@misc{semantic:gan:jacob2022steexsteeringcounterfactualexplanations,
      title={STEEX: Steering Counterfactual Explanations with Semantics}, 
      author={Paul Jacob and Éloi Zablocki and Hédi Ben-Younes and Mickaël Chen and Patrick Pérez and Matthieu Cord},
      year={2022},
      eprint={2111.09094},
      archivePrefix={arXiv},
      primaryClass={cs.CV},
      url={https://arxiv.org/abs/2111.09094}, 
}

@misc{semantic:gan:samadi2023safesaliencyawarecounterfactualexplanations,
      title={SAFE: Saliency-Aware Counterfactual Explanations for DNN-based Automated Driving Systems}, 
      author={Amir Samadi and Amir Shirian and Konstantinos Koufos and Kurt Debattista and Mehrdad Dianati},
      year={2023},
      eprint={2307.15786},
      archivePrefix={arXiv},
      primaryClass={cs.LG},
      url={https://arxiv.org/abs/2307.15786}, 
}

@misc{cycle:gan:ghandeharioun2022dissectdisentangledsimultaneousexplanations,
      title={DISSECT: Disentangled Simultaneous Explanations via Concept Traversals}, 
      author={Asma Ghandeharioun and Been Kim and Chun-Liang Li and Brendan Jou and Brian Eoff and Rosalind W. Picard},
      year={2022},
      eprint={2105.15164},
      archivePrefix={arXiv},
      primaryClass={cs.LG},
      url={https://arxiv.org/abs/2105.15164}, 
}

@misc{cycle:gan:khorram2022cycleconsistentcounterfactualslatenttransformations,
      title={Cycle-Consistent Counterfactuals by Latent Transformations}, 
      author={Saeed Khorram and Li Fuxin},
      year={2022},
      eprint={2203.15064},
      archivePrefix={arXiv},
      primaryClass={cs.CV},
      url={https://arxiv.org/abs/2203.15064}, 
}

@misc{style:gan:lang2021explainingstyletraininggan,
      title={Explaining in Style: Training a GAN to explain a classifier in StyleSpace}, 
      author={Oran Lang and Yossi Gandelsman and Michal Yarom and Yoav Wald and Gal Elidan and Avinatan Hassidim and William T. Freeman and Phillip Isola and Amir Globerson and Michal Irani and Inbar Mosseri},
      year={2021},
      eprint={2104.13369},
      archivePrefix={arXiv},
      primaryClass={cs.CV},
      url={https://arxiv.org/abs/2104.13369}, 
}

@misc{adamw:loshchilov2019decoupledweightdecayregularization,
      title={Decoupled Weight Decay Regularization}, 
      author={Ilya Loshchilov and Frank Hutter},
      year={2019},
      eprint={1711.05101},
      archivePrefix={arXiv},
      primaryClass={cs.LG},
      url={https://arxiv.org/abs/1711.05101}, 
}

@misc{cosine:loshchilov2017sgdrstochasticgradientdescent,
      title={SGDR: Stochastic Gradient Descent with Warm Restarts}, 
      author={Ilya Loshchilov and Frank Hutter},
      year={2017},
      eprint={1608.03983},
      archivePrefix={arXiv},
      primaryClass={cs.LG},
      url={https://arxiv.org/abs/1608.03983}, 
}

@inproceedings{CelebA:liu2015faceattributes,
  title = {Deep Learning Face Attributes in the Wild},
  author = {Liu, Ziwei and Luo, Ping and Wang, Xiaogang and Tang, Xiaoou},
  booktitle = {Proceedings of International Conference on Computer Vision (ICCV)},
  month = {December},
  year = {2015} 
}

@misc{unet:ronneberger2015unetconvolutionalnetworksbiomedical,
      title={U-Net: Convolutional Networks for Biomedical Image Segmentation}, 
      author={Olaf Ronneberger and Philipp Fischer and Thomas Brox},
      year={2015},
      eprint={1505.04597},
      archivePrefix={arXiv},
      primaryClass={cs.CV},
      url={https://arxiv.org/abs/1505.04597}, 
}

@misc{CBM:koh2020conceptbottleneckmodels,
      title={Concept Bottleneck Models}, 
      author={Pang Wei Koh and Thao Nguyen and Yew Siang Tang and Stephen Mussmann and Emma Pierson and Been Kim and Percy Liang},
      year={2020},
      eprint={2007.04612},
      archivePrefix={arXiv},
      primaryClass={cs.LG},
      url={https://arxiv.org/abs/2007.04612}, 
}

@inproceedings{CBM:DDPM:ismail2024concept,
title={Concept Bottleneck Generative Models},
author={Aya Abdelsalam Ismail and Julius Adebayo and Hector Corrada Bravo and Stephen Ra and Kyunghyun Cho},
booktitle={The Twelfth International Conference on Learning Representations},
year={2024},
url={https://openreview.net/forum?id=L9U5MJJleF}
}

@misc{CFE:2:augustin2022diffusionvisualcounterfactualexplanations,
      title={Diffusion Visual Counterfactual Explanations}, 
      author={Maximilian Augustin and Valentyn Boreiko and Francesco Croce and Matthias Hein},
      year={2022},
      eprint={2210.11841},
      archivePrefix={arXiv},
      primaryClass={cs.CV},
      url={https://arxiv.org/abs/2210.11841}, 
}

@inproceedings{CFE:1:jeanneret2022diffusion,
  title={Diffusion models for counterfactual explanations},
  author={Jeanneret, Guillaume and Simon, Lo{\"\i}c and Jurie, Fr{\'e}d{\'e}ric},
  booktitle={Proceedings of the Asian conference on computer vision},
  pages={858--876},
  year={2022}
}

@article{score:song2020score,
  title={Score-based generative modeling through stochastic differential equations},
  author={Song, Yang and Sohl-Dickstein, Jascha and Kingma, Diederik P and Kumar, Abhishek and Ermon, Stefano and Poole, Ben},
  journal={arXiv preprint arXiv:2011.13456},
  year={2020}
}

@misc{DDPM:classifier:dhariwal2021diffusionmodelsbeatgans,
      title={Diffusion Models Beat GANs on Image Synthesis}, 
      author={Prafulla Dhariwal and Alex Nichol},
      year={2021},
      eprint={2105.05233},
      archivePrefix={arXiv},
      primaryClass={cs.LG},
      url={https://arxiv.org/abs/2105.05233}, 
}

@misc{DDPM:classifier-free:ho2022classifierfreediffusionguidance,
      title={Classifier-Free Diffusion Guidance}, 
      author={Jonathan Ho and Tim Salimans},
      year={2022},
      eprint={2207.12598},
      archivePrefix={arXiv},
      primaryClass={cs.LG},
      url={https://arxiv.org/abs/2207.12598}, 
}

@article{Selvaraju_2019,
   title={Grad-CAM: Visual Explanations from Deep Networks via Gradient-Based Localization},
   volume={128},
   ISSN={1573-1405},
   url={http://dx.doi.org/10.1007/s11263-019-01228-7},
   DOI={10.1007/s11263-019-01228-7},
   number={2},
   journal={International Journal of Computer Vision},
   publisher={Springer Science and Business Media LLC},
   author={Selvaraju, Ramprasaath R. and Cogswell, Michael and Das, Abhishek and Vedantam, Ramakrishna and Parikh, Devi and Batra, Dhruv},
   year={2019},
   month=oct, pages={336–359} }

@article{medical:tjoa2020survey,
  title={A survey on explainable artificial intelligence (xai): Toward medical xai},
  author={Tjoa, Erico and Guan, Cuntai},
  journal={IEEE transactions on neural networks and learning systems},
  volume={32},
  number={11},
  pages={4793--4813},
  year={2020},
  publisher={IEEE}
}

@article{medical:krishnan2022self,
  title={Self-supervised learning in medicine and healthcare},
  author={Krishnan, Rayan and Rajpurkar, Pranav and Topol, Eric J},
  journal={Nature Biomedical Engineering},
  volume={6},
  number={12},
  pages={1346--1352},
  year={2022},
  publisher={Nature Publishing Group UK London}
}

@article{car:atakishiyev2024explainable,
  title={Explainable artificial intelligence for autonomous driving: A comprehensive overview and field guide for future research directions},
  author={Atakishiyev, Shahin and Salameh, Mohammad and Yao, Hengshuai and Goebel, Randy},
  journal={IEEE Access},
  year={2024},
  publisher={IEEE}
}

@article{blackbox:rai2020explainable,
  title={Explainable AI: From black box to glass box},
  author={Rai, Arun},
  journal={Journal of the academy of marketing science},
  volume={48},
  number={1},
  pages={137--141},
  year={2020},
  publisher={Springer}
}

@article{explain:dwivedi2023explainable,
  title={Explainable AI (XAI): Core ideas, techniques, and solutions},
  author={Dwivedi, Rudresh and Dave, Devam and Naik, Het and Singhal, Smiti and Omer, Rana and Patel, Pankesh and Qian, Bin and Wen, Zhenyu and Shah, Tejal and Morgan, Graham and others},
  journal={ACM computing surveys},
  volume={55},
  number={9},
  pages={1--33},
  year={2023},
  publisher={ACM New York, NY}
}

@inproceedings{VCE:goyal2019counterfactual,
  title={Counterfactual visual explanations},
  author={Goyal, Yash and Wu, Ziyan and Ernst, Jan and Batra, Dhruv and Parikh, Devi and Lee, Stefan},
  booktitle={International Conference on Machine Learning},
  pages={2376--2384},
  year={2019},
  organization={PMLR}
}

@article{attribusion:method:adebayo2018sanity,
  title={Sanity checks for saliency maps},
  author={Adebayo, Julius and Gilmer, Justin and Muelly, Michael and Goodfellow, Ian and Hardt, Moritz and Kim, Been},
  journal={Advances in neural information processing systems},
  volume={31},
  year={2018}
}

@article{GAN:VCE:medical:mertes2022ganterfactual,
  title={Ganterfactual—counterfactual explanations for medical non-experts using generative adversarial learning},
  author={Mertes, Silvan and Huber, Tobias and Weitz, Katharina and Heimerl, Alexander and Andr{\'e}, Elisabeth},
  journal={Frontiers in artificial intelligence},
  volume={5},
  pages={825565},
  year={2022},
  publisher={Frontiers Media SA}
}

@article{AE:VCE:freiesleben2022intriguing,
  title={The intriguing relation between counterfactual explanations and adversarial examples},
  author={Freiesleben, Timo},
  journal={Minds and Machines},
  volume={32},
  number={1},
  pages={77--109},
  year={2022},
  publisher={Springer}
}

@article{cond:ddpm:panagiotakopoulos2025conditional,
  title={Conditional Diffusion Models: A Survey of Techniques, Applications and Challenges},
  author={Panagiotakopoulos, Theodor and Kotsiantis, Sotiris and Gkillas, Alexandros and Lalos, Aris S},
  journal={IEEE Access},
  year={2025},
  publisher={IEEE}
}

@InProceedings{CFE:ADVER:Jeanneret_2023_CVPR,
    author    = {Jeanneret, Guillaume and Simon, Lo{\"\i}c and Jurie, Fr\'ed\'eric},
    title     = {Adversarial Counterfactual Visual Explanations},
    booktitle = {Proceedings of the IEEE/CVF Conference on Computer Vision and Pattern Recognition (CVPR)},
    month     = {June},
    year      = {2023},
    pages     = {16425-16435}
}

@misc{dominici2025counterfactualconceptbottleneckmodels,
      title={Counterfactual Concept Bottleneck Models}, 
      author={Gabriele Dominici and Pietro Barbiero and Francesco Giannini and Martin Gjoreski and Giuseppe Marra and Marc Langheinrich},
      year={2025},
      eprint={2402.01408},
      archivePrefix={arXiv},
      primaryClass={cs.LG},
      url={https://arxiv.org/abs/2402.01408}, 
}

@misc{VCE:NO:GENERATIVEMODEL:chang2021robustclassificationmodelcounterfactual,
      title={Towards Robust Classification Model by Counterfactual and Invariant Data Generation}, 
      author={Chun-Hao Chang and George Alexandru Adam and Anna Goldenberg},
      year={2021},
      eprint={2106.01127},
      archivePrefix={arXiv},
      primaryClass={cs.CV},
      url={https://arxiv.org/abs/2106.01127}, 
}

@misc{CF:FAST:weng2024fastdiffusionbasedcounterfactualsshortcut,
      title={Fast Diffusion-Based Counterfactuals for Shortcut Removal and Generation}, 
      author={Nina Weng and Paraskevas Pegios and Eike Petersen and Aasa Feragen and Siavash Bigdeli},
      year={2024},
      eprint={2312.14223},
      archivePrefix={arXiv},
      primaryClass={cs.CV},
      url={https://arxiv.org/abs/2312.14223}, 
}

@inproceedings{CF:LATENT:DDPM:Luu_2025,
   title={From Visual Explanations to Counterfactual Explanations with Latent Diffusion},
   url={http://dx.doi.org/10.1109/WACV61041.2025.00051},
   DOI={10.1109/wacv61041.2025.00051},
   booktitle={2025 IEEE/CVF Winter Conference on Applications of Computer Vision (WACV)},
   publisher={IEEE},
   author={Luu, Tung and Le, Nam and Le, Duc and Le, Bac},
   year={2025},
   month=feb, pages={420–429} }

@misc{CF:COCO:le2023cococounterfactualsautomaticallyconstructedcounterfactual,
      title={COCO-Counterfactuals: Automatically Constructed Counterfactual Examples for Image-Text Pairs}, 
      author={Tiep Le and Vasudev Lal and Phillip Howard},
      year={2023},
      eprint={2309.14356},
      archivePrefix={arXiv},
      primaryClass={cs.LG},
      url={https://arxiv.org/abs/2309.14356}, 
}

@article{DEEP:FACE:serengil2024lightface,
  title     = {A Benchmark of Facial Recognition Pipelines and Co-Usability Performances of Modules},
  author    = {Serengil, Sefik and Ozpinar, Alper},
  journal   = {Journal of Information Technologies},
  volume    = {17},
  number    = {2},
  pages     = {95-107},
  year      = {2024},
  doi       = {10.17671/gazibtd.1399077},
  url       = {https://dergipark.org.tr/en/pub/gazibtd/issue/84331/1399077},
  publisher = {Gazi University}
}

@article{dominici2024causal,
  title={Causal concept graph models: Beyond causal opacity in deep learning},
  author={Dominici, Gabriele and Barbiero, Pietro and Zarlenga, Mateo Espinosa and Termine, Alberto and Gjoreski, Martin and Marra, Giuseppe and Langheinrich, Marc},
  journal={arXiv preprint arXiv:2405.16507},
  year={2024}
}

@misc{stable:esser2024scalingrectifiedflowtransformers,
      title={Scaling Rectified Flow Transformers for High-Resolution Image Synthesis}, 
      author={Patrick Esser and Sumith Kulal and Andreas Blattmann and Rahim Entezari and Jonas Müller and Harry Saini and Yam Levi and Dominik Lorenz and Axel Sauer and Frederic Boesel and Dustin Podell and Tim Dockhorn and Zion English and Kyle Lacey and Alex Goodwin and Yannik Marek and Robin Rombach},
      year={2024},
      eprint={2403.03206},
      archivePrefix={arXiv},
      primaryClass={cs.CV},
      url={https://arxiv.org/abs/2403.03206}, 
}

\end{document}